\documentclass[runningheads]{llncs}

\usepackage[mobile]{eccv}

\usepackage{eccvabbrv}

\usepackage{graphicx}
\usepackage{booktabs}
\usepackage{siunitx}

\usepackage[accsupp]{axessibility}  

\usepackage{colortbl} 
\usepackage[dvipsnames]{xcolor}

\usepackage[pagebackref,colorlinks]{hyperref}

\usepackage{orcidlink}

\definecolor{TableDarkGreen}{RGB}{182,215,168}
\definecolor{TableLightGreen}{RGB}{207,234,215}
\definecolor{TableYellow}{RGB}{255,242,204}
\definecolor{TableRed}{RGB}{244,204,204}

\newcommand{\pose}{T}
\newcommand{\crd}{\mathbf{y}}

\newcommand{\pos}{\mathbf{p}}
\newcommand{\proj}{\boldsymbol{\pi}}

\newcommand{\colmapreloc}{``Sparse COLMAP + Reloc + BA''\xspace}

\usepackage{array}
\usepackage{multirow, makecell}
\usepackage{arydshln}
\usepackage{hhline}
\usepackage[most]{tcolorbox} 
\usepackage[export]{adjustbox}

\newcommand{\TCBTableYellow}[1]{\tcbox[on line,boxsep=0pt,left=0pt,right=0pt,top=0pt,bottom=0pt,colframe=white,colback=TableYellow,highlight math style={enhanced},extrude by=0.75pt]{#1}}
\newcommand{\TCBTableDarkGreen}[1]{\tcbox[on line,boxsep=0pt,left=0pt,right=0pt,top=0pt,bottom=0pt,colframe=white,colback=TableDarkGreen,highlight math style={enhanced},extrude by=0.75pt]{#1}}
\newcommand{\TCBTableLightGreen}[1]{\tcbox[on line,boxsep=0pt,left=0pt,right=0pt,top=0pt,bottom=0pt,colframe=white,colback=TableLightGreen,highlight math style={enhanced},extrude by=0.75pt]{#1}}
\newcommand{\TCBTableRed}[1]{\tcbox[on line,boxsep=0pt,left=0pt,right=0pt,top=0pt,bottom=0pt,colframe=white,colback=TableRed,highlight math style={enhanced},extrude by=0.75pt]{#1}}

\DeclareCaptionSubType*{table}
\begin{document}

\title{Scene Coordinate Reconstruction:\\ Posing of Image Collections via Incremental Learning of a Relocalizer} 

\titlerunning{Scene Coordinate Reconstruction}

\author{%
Eric Brachmann\inst{1} \and
Jamie Wynn\inst{1} \and
Shuai Chen\inst{2} \and
Tommaso Cavallari\inst{1} \and
\mbox{{\'{A}}ron Monszpart\inst{1}} \and
Daniyar Turmukhambetov\inst{1} \and
Victor Adrian Prisacariu\inst{1,2}%
}

\authorrunning{E. Brachmann et al.}

\institute{%
Niantic \and University of Oxford%
}

\maketitle

\begin{abstract}
We address the task of estimating camera parameters from a set of images depicting a scene. 
Popular feature-based structure-from-motion (SfM) tools solve this task by incremental reconstruction: 
they repeat triangulation of sparse 3D points and registration of more camera views to the sparse point cloud. 
We re-interpret incremental structure-from-motion as an iterated application and refinement of a visual relocalizer, that is, of a method that registers new views to the current state of the reconstruction. 
This perspective allows us to investigate alternative visual relocalizers that are not rooted in local feature matching. 
We show that \emph{scene coordinate regression}, a learning-based relocalization approach, allows us to build implicit, neural scene representations from unposed images.
Different from other learning-based reconstruction methods, we do not require pose priors nor sequential inputs, and we optimize efficiently over thousands of images.
In many cases, our method, ACE0, estimates camera poses with an accuracy close to  feature-based SfM, as demonstrated by novel view synthesis.\\
Project page: \url{https://nianticlabs.github.io/acezero/}
\end{abstract}

\section{Introduction}
\label{sec:intro}

\begin{figure*}[t]
  \centering
   \includegraphics[width=1.0\linewidth]{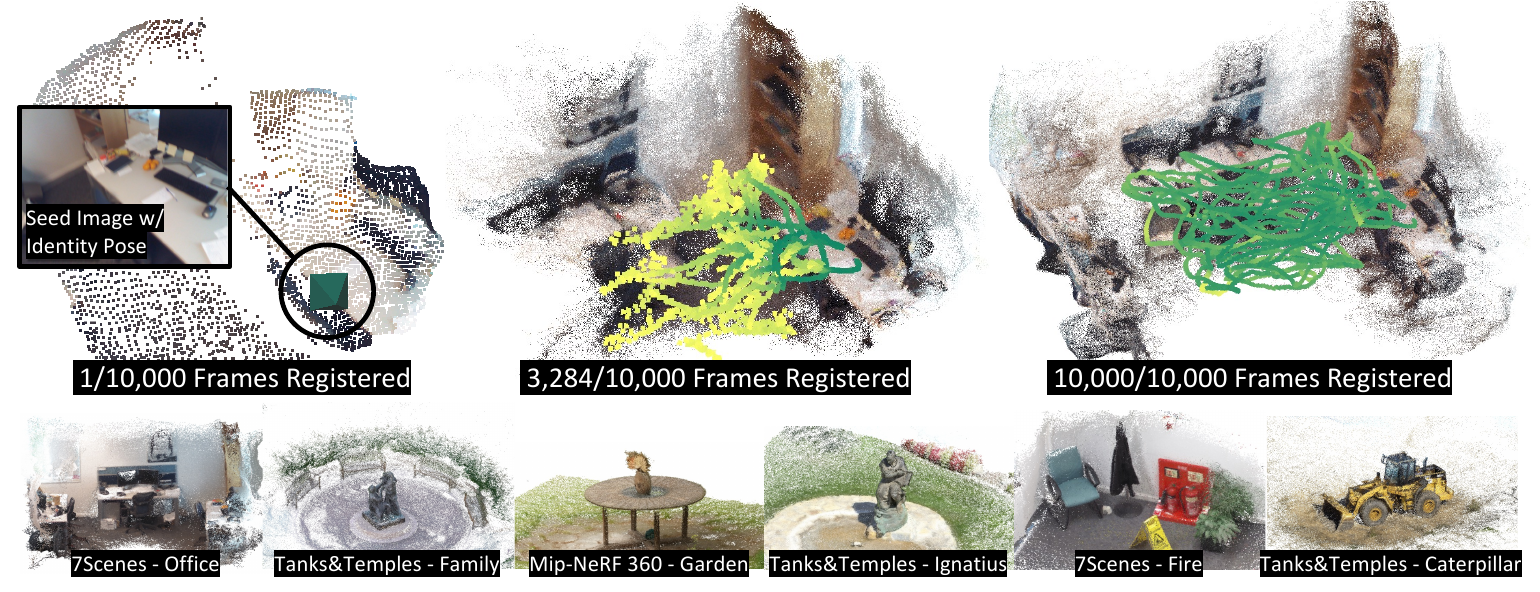}
   \caption{\textbf{Reconstructing 10,000 Images. Top:} Starting from a single image and the identity pose, we train a learning-based visual relocalizer. The relocalizer allows us to estimate the poses of more views, and the additional views allow us to refine the relocalizer. We show three out of six iterations for this scene (7Scenes Office~\cite{shotton2013scene}). 
   All 10k images have been posed in roughly 1 hour on a single GPU. In comparison NoPe-NeRF~\cite{bian2023nope} needs two days to pose 200 images.
   The point cloud is a visualization of the implicit scene representation of the relocalizer. 
   Camera positions are color coded by relocalization confidence from \TCBTableYellow{yellow} (low) to \TCBTableDarkGreen{green} (high). \textbf{Bottom:} Point clouds from Nerfacto~\cite{nerfstudio} trained on top of our poses for a few scenes from our experiments.}
   \label{fig:teaser}
\end{figure*}

\begin{center}
  \emph{In the beginning there was structure-from-motion.}
\end{center}
The genesis of numerous computer vision tasks lies in the estimation of camera poses and scene geometry from a set of images. 
It is the first fundamental step that lets us leave the image plane and venture into 3D. 
Since structure-from-motion (SfM) is such a central capability, we have researched it for decades. 
By now, refined open-source tools, such as COLMAP~\cite{schoenberger2016sfm}, and efficient commercial packages, such as RealityCapture~\cite{reality-capture}, are available to us.
Feature matching-based SfM is the gold standard for estimating poses from images,
with a precision that makes its estimates occasionally considered ``ground truth''~\cite{brachmann2021limits,jin2021image,reizenstein2021common,arnold2022map,barron2022mipnerf360}. 

The success of Neural Radiance Fields (NeRF)~\cite{mildenhall2020nerf} has renewed interest in the question of whether SfM can be solved differently, based on neural, implicit scene representations rather than 3D point clouds.
There has been some progress in recent years but, thus far, learning-based approaches to camera pose recovery still have significant limitations.
They either require coarse initial poses \cite{jeong2021self,wang2021nerf,lin2021barf,xia2022sinerf}, prior knowledge of the pose distribution~\cite{meng2021gnerf} or sequential inputs~\cite{Bloesch2018codeslam,teed2021droid,bian2023nope}.
In terms of the number of images that learning-based approaches can handle, they are either explicitly targeted at few-frame problems \cite{ummenhofer2017demon,wei2020deepsfm,sinha2023sparsepose,lin2024relposepp,zhang2024raydiffusion,wang2024dust3r} or they are computationally so demanding that they can realistically only be applied to a few hundred images at most \cite{wang2021nerf,lin2021barf,bian2023nope}.
We show that none of these limitations are an inherent consequence of using learning-based scene representations.

Our approach is inspired by incremental SfM and its relationship to another computer vision task: visual relocalization. 
Visual relocalization describes the problem of estimating the camera pose of a query image \wrt to an existing scene map. 
Incremental SfM can be re-interpreted as a loop of 1) do visual relocalization to register new views to the reconstruction, and 2) refine/extend the reconstruction using the newly registered views.
Local feature matching is a traditional approach to visual relocalization~\cite{sattler2012improving,sattler2016efficient,sattler2017large,sarlin2019coarse}.
In recent years, multiple learning-based relocalizers have been proposed that encode the scene implicitly in the weights of a neural network~\cite{kendall2015posenet,arnold2022map,brachmann2017dsac,cavallari2019let,brachmann2018dsacpp,brachmann2021dsacstar,brachmann2023accelerated}. 
Not all of them are suitable for building a SfM pipeline. 
We need a relocalizer with high accuracy and good generalization. Training of the relocalizer has to be swift. 
We need to be able to bootstrap relocalization without ground truth poses. 
And we need to be able to tell whether registration of a new image was successful. 

We show that ``scene coordinate regression'', an approach to visual relocalization proposed a decade ago~\cite{shotton2013scene}, has the desirable properties and can serve as the core of a new approach to learning-based SfM: Scene coordinate \emph{reconstruction}.
Rather than optimizing over image-to-image matches, like feature-based SfM, scene coordinate reconstruction regresses image-to-\emph{scene} correspondences directly.
Rather than representing the scene as a 3D point cloud with high dimensional descriptors, we encode the scene into a lightweight neural network.
Our approach works on unsorted images without pose priors and efficiently optimises over thousands of images, see \Cref{fig:teaser}.

\noindent We summarize our \textbf{contributions}:
\begin{itemize}
    \item \emph{Scene Coordinate Reconstruction}, a new approach to SfM based on incremental learning of scene coordinate regression, a visual relocalization principle.
    \item We turn the fast-learning visual relocalizer ACE~\cite{brachmann2023accelerated} into a SfM framework that is able to predict the camera poses of a set of unposed RGB images. We refer to this new SfM pipeline as \emph{ACE0} (ACE Zero).
    \item Compared to ACE~\cite{brachmann2023accelerated}, we add the capability to train in a self-supervised fashion. We start from a single image, and iterate between learning the map and registering new views. We expedite reconstruction times by early stopping, and increase reconstruction quality by pose refinement.
\end{itemize}

\section{Related Work}

\paragraph{Reconstruction.}
SfM pipelines either ingest a collection of \emph{unordered images} ~\cite{snavely2006photo,brown2005unsupervised,schaffalitzky2002multi} or \emph{an ordered image sequence}~\cite{szeliski1994recovering,beardsley1997sequential,pollefeys2000automated,nister2004visual,davison2003real} from a video to recover 3D structure of a scene and camera poses (``motion'') of the images.

SIFT~\cite{lowe2004distinctive} and other robust descriptors allow matching image features across wide baselines enabling systems to reconstruct large-scale scenes using Internet images~\cite{snavely2006photo,snavely2008modeling,wu2013towards,schoenberger2016sfm}.
Image-to-image matches can also be regressed directly in a detector-free setup \cite{sun2021loftr, he2024detectorfree}.
Feature tracks across multiple images are built from image-to-image matches.
Feature tracks and estimated relative poses are used to solve for the 3D feature coordinates, camera poses and calibrations (intrinsic matrices).
This geometric optimization problem is mainly solved using bundle adjustment which was explored in photogrammetry and geodesy~\cite{brown1976bundle,kraus1993photogrammetry} and became standard in the computer vision community~\cite{triggs2000bundle,hartley2003multiple}.
Bundle adjustment relies on the initialization being close to the solution (\ie, camera poses and 3D points are already mostly accurate).

There are two main approaches to this problem.
\emph{Incremental SfM}~\cite{beardsley1997sequential,pollefeys2000automated} starts the reconstruction from very few images to create a high-quality seed reconstruction that progressively grows by registering more images and refining the reconstruction until convergence.
\emph{Global SfM} approaches solve for ``global'' poses of all images using estimates of relative poses, \ie, motion averaging~\cite{govindu2001combining,govindu2004lie}, rotation averaging~\cite{hartley2013rotation,martinec2007robust} and pose-graph optimization~\cite{carlone2015initialization}.
Various techniques were proposed to improve SfM runtime for very large sets of images~\cite{crandall2011discrete,agarwal2010bundle,agarwal2011building,heinly2015reconstructing,bhowmick2015divide,bhowmick2017divide,snavely2008skeletal,gherardi2010improving,toldo2015hierarchical}.
Our work is similar to Incremental SfM, as we also progressively register images to the reconstructed scene starting from a seed reconstruction.
However, we do not explicitly compute image matches, nor feature tracks across images, which can be computationally expensive.

\paragraph{Visual Relocalization.}

A reconstructed (or mapped) scene is a database of images with known camera poses. This database can be used by a visual relocalizer to estimate poses for new query images to ``relocalize'' a camera in the scene. Feature-based approaches extract 2D local features~\cite{lowe2004distinctive,detone2018superpoint,dusmanu2019d2,sarlin2020superglue,lindenberger2023lightglue,sun2021loftr,sarlin2021back} from a query image and match them to 3D points to solve for the query pose using perspective-n-point  (PnP) solvers~\cite{pnp}, \mbox{\eg,~\cite{sattler2011fast,sattler2012improving,sattler2016efficient}}; or match query features to 2D local features of mapped images to triangulate the query image, \eg,~\cite{zhang2006image,zhou2020learn}. For large scenes, matching features on a subset of database images relevant to the query can improve the speed and accuracy, \eg,~\cite{sattler2017large,sarlin2019coarse,rau2020predicting,humenberger2022investigating}. 

Some learning-based approaches encode the map of a scene in the weights of a neural network. PoseNet~\cite{kendall2015posenet} directly regresses the absolute camera pose given an input image using a CNN that was trained on image-pose pairs of the reconstructed scene. Sattler \etal show that extrapolating outside the mapped poses can be challenging for absolute pose regression approaches~\cite{sattler2019understanding}. Relative pose regression networks~\cite{balntas2018relocnet,ding2019camnet,laskar2017camera,turkoglu2021visual} estimate the relative pose for a pair of images, allowing triangulation of the query image from multiple map images, or estimate a scale-metric relative pose \wrt a single map image \cite{arnold2022map}.
Scene coordinate regression~\cite{shotton2013scene,cavallari2017fly,brachmann2017dsac,brachmann2018dsacpp,cavallari2019let,brachmann2021dsacstar,brachmann2023accelerated,li2020hierarchical} directly predicts 3D coordinates of a point given a patch. This approach generalizes well as the camera pose is solved using PnP~\cite{pnp} within a RANSAC loop~\cite{ransac}. ACE~\cite{brachmann2023accelerated} shows that training of scene coordinate regression can be greatly accelerated.
In our work we train the ACE localizer, but we start from images without poses.

\paragraph{Image-Based Rendering.}
In recent years, neural radiance fields (NeRFs)~\cite{mildenhall2020nerf} have seen a lot of attention from the community. NeRFs allow photorealistic novel view synthesis when trained from image-pose pairs. Typically, estimation of the image poses is done in advance by using an SfM pipeline such as COLMAP, \eg,~\cite{barron2022mipnerf360}.
Nonetheless, research exists that estimates camera poses with NeRFs, facilitates camera localization for novel views after NeRF training~\cite{yen2021inerf,chen21,chen2022dfnet,chen2023refinement,moreau2023crossfire}, or simultaneously estimates camera poses during NeRF training from images alone~\cite{wang2021nerf,lin2021barf,bian2023nope,sinha2023sparsepose,cheng2023lunerf}. However, these approaches either assume that the scene is captured from the front~\cite{jeong2021self,wang2021nerf,xia2022sinerf}, that coarse poses are available for initialization~\cite{lin2021barf}, or that images are captured sequentially~\cite{bian2023nope}. Techniques that rely on a multi-layer perceptron (MLP) representation of the scene, \eg~\cite{bian2023nope,cheng2023lunerf}, are slow to train, taking days to converge. While radiance fields can be trained faster using multi-layer hash grids~\cite{mueller2022instant} or Gaussian splats~\cite{kerbl3Dgaussians}, their efficacy in pose estimation without approximate prior pose initialization~\cite{lin2023icrapnerf,moreau2023crossfire} or sequential image inputs~\cite{kerbl3Dgaussians} remains unproven. Concurrently, several learning-based camera pose estimation methods have been proposed~\cite{zhang2024raydiffusion,wang2024dust3r,lin2024relposepp}. Due to GPU memory constraints, these methods estimate poses for sparse sets of images.

\section{Method}

\begin{figure*}[t]
  \centering
   \includegraphics[width=1.0\linewidth]{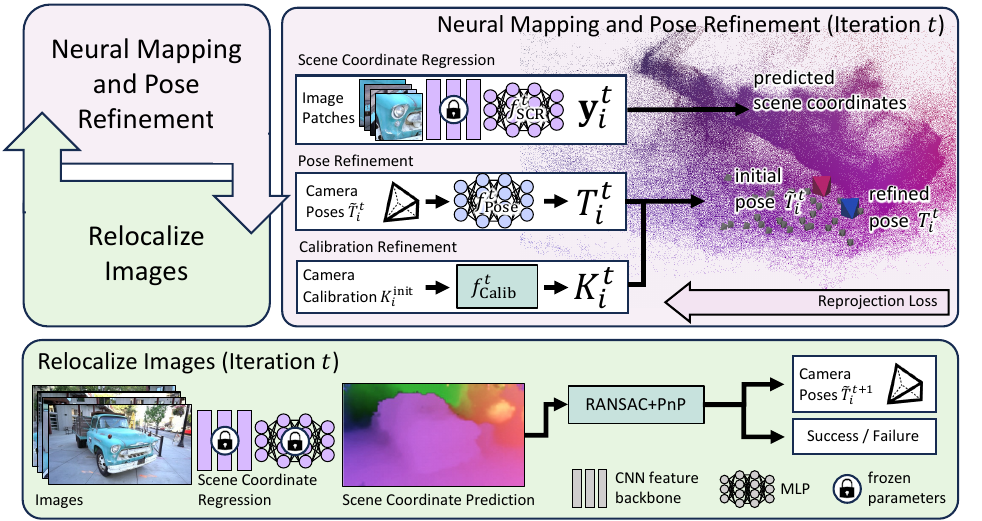}
   \caption{\textbf{ACE0 Framework. }\textbf{Top left:} We loop between learning a reconstruction from the current set of images and poses (``neural mapping''), and estimating poses of more images (``relocalization''). \textbf{Top right:} During the mapping stage, we train a scene coordinate regression network as our scene representation. Camera poses of the last relocalization round and camera calibration parameters are refined during this process. We visualize scene coordinates by mapping XYZ to the RGB cube. \textbf{Bottom:} In the relocalization stage, we re-estimate poses of images using the scene coordinate regression network, including images that were previously not registered to the reconstruction. If the registration of an image succeeds, it will be used in the next iteration of the mapping stage; otherwise it will not.}
   \label{fig:system_ace0}
\end{figure*}

\noindent \textbf{Preliminaries.} 
The input to our system is a set of RGB images, denoted by $\mathcal{I} = \{I_i\}$, where $i$ refers to the image index.
Our system estimates the corresponding set of camera parameters, both intrinsics and extrinsics: $\mathcal{H} = \{(K_i, \pose_i)\}$.
Each $\pose_i$ refers to a $3\times4$ matrix containing a rotation and translation, while $K_i$ refers to a $3\times3$ matrix with the calibration parameters.
We assume no particular image order or any prior knowledge about the pose distribution.

We also want to recover the 3D structure of the scene: 
Each pixel $j$ in image $i$ with 2D pixel position $\pos_{ij}$ has a corresponding coordinate in 3D, denoted as $\crd_{ij}$.
The 2D pixel positions and 3D scene coordinates are related by the camera pose and the projection function $\proj$:
\begin{equation}
    \pos_{ij} = \proj(K_i, \pose_i, \crd_{ij}),
\end{equation}
where $\pose_i$ maps camera coordinates to scene coordinates, and $K_i$ projects camera coordinates to the image plane.

As our scene representation, we utilize a scene coordinate regression model~\cite{shotton2013scene}, \ie, a learnable function $f_\text{SCR}$ that maps an image patch of image $I_i$, centered around pixel position $\pos_{ij}$ to a scene coordinate: $\crd_{ij} = f_\text{SCR}(\pos_{ij}, I_i)$.

Given a set of 2D-3D correspondences predicted by $f_\text{SCR}$ for any image $I_i$, we can recover this image's camera pose $\pose_i$ using a pose solver $g$:
\begin{equation}
    \pose_i = g\left(K_i, \{(\pos_{ij}, \crd_{ij})\}\right).
\end{equation}
Since 2D-3D correspondences can be inaccurate, and contain incorrect predictions, $g$ combines a PnP solver~\cite{pnp} with a RANSAC loop~\cite{ransac}.

Normally, scene coordinate regression models are trained in a supervised fashion for the task of visual relocalization~\cite{shotton2013scene,cavallari2017fly,brachmann2017dsac,cavallari2019let,brachmann2023accelerated, brachmann2021dsacstar,brachmann2018dsacpp,li2020hierarchical}. 
That is, $f_\text{SCR}$ is trained using images with known ground truth camera parameters $\{(I_i, \pose_i^\text{GT}, K_i^\text{GT})\}$, and used for estimating the poses of unseen query images.
Instead, we show how these models can be trained self-supervised, without ground truth poses, to estimate the poses of the mapping images themselves.
Thus, we turn scene coordinate regression into scene coordinate reconstruction, a learning-based SfM tool.

\subsection{Neural Mapping}

We train the scene coordinate regression model iteratively where we denote the current time step as $t$ and the corresponding scene model as $f^t_\text{SCR}$. 
We iterate between training the scene model, and registering new views, see \Cref{fig:system_ace0}.

At iteration $t$ we assume that a subset of images has already been registered to the scene, $\mathcal{I}^t_\text{Reg} \subset \mathcal{I}$, and where corresponding camera parameters, $\pose^t_i$ and $K^t_i$, have already been estimated.
Using these as pseudo ground truth, we train the scene model by minimizing the pixel-wise reprojection error:
\begin{equation}
    \sum_{I_i \in \mathcal{I}^t_\text{Reg}} \sum_{j \in I_i} \left\Vert \pos_{ij} - \proj(K_i^t, \pose_i^t, \crd_{ij}^t) \right\Vert,
    \label{eq:repro_objective}
\end{equation}
where the scene model $f^t_\text{SCR}$ predicts coordinates $\crd_{ij}^t$.

\noindent \textbf{Mapping Framework.}
We optimize Eq.~\ref{eq:repro_objective} using stochastic gradient descent, using the fast-learning scene coordinate regressor ACE~\cite{brachmann2023accelerated} (Accelerated Coordinate Encoding). 
ACE is trained in minutes, even for thousands of views.
The training speed is important since we have to train the model in multiple iterations.
ACE approximates Eq.~\ref{eq:repro_objective} by sampling random patches from the training set, $\mathcal{I}^t_\text{Reg}$. 
To do that efficiently, ACE employs a pre-trained encoder to pre-compute high dimensional features for a large number of training patches. 
The encoder stays frozen during mapping. 
The actual scene model, which is trained, is a multi-layer perceptron (MLP) that maps encoder features to scene coordinates; see Fig.~\ref{fig:system_ace0} (top right) for a visual representation.

\begin{figure*}[t]
  \centering
   \includegraphics[width=1.0\linewidth]{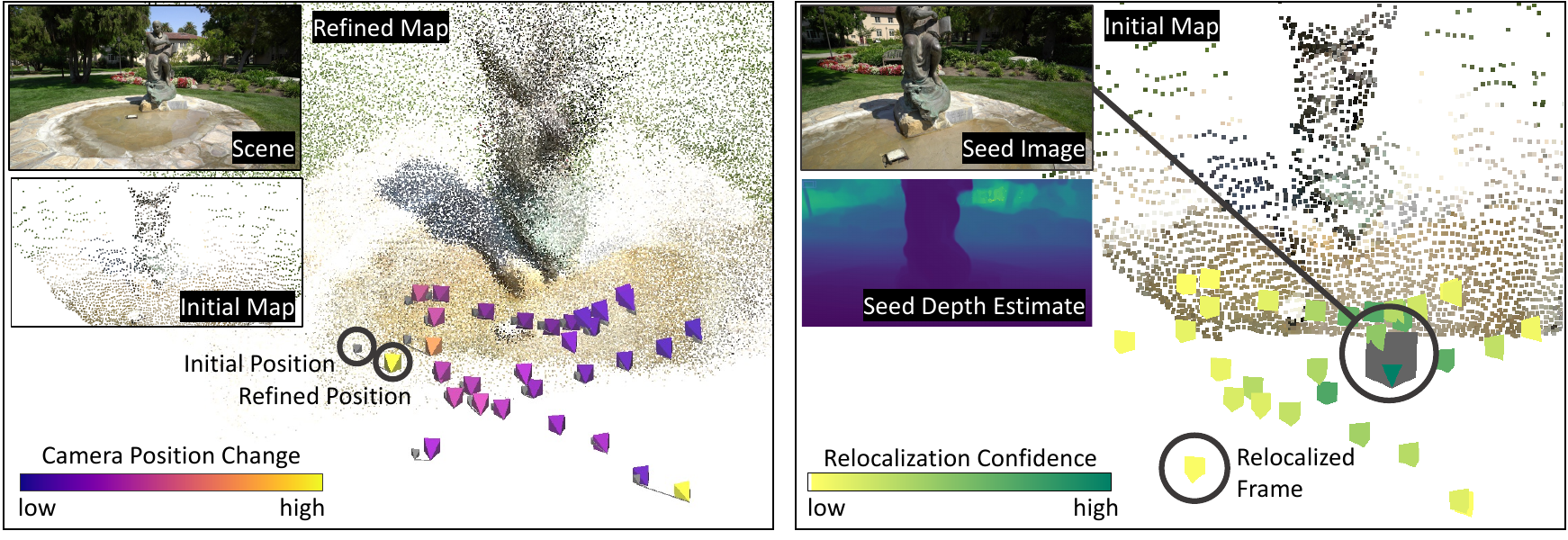}
   \caption{\textbf{Left: Pose Refinement.} Since we register images based on a coarse and incomplete state of the reconstruction, we add the ability to refine poses during neural mapping. An MLP predicts pose updates relative to the initial poses, supervised by the reprojection error of scene coordinates. \textbf{Right: Initialization.} To start the reconstruction, we train the network using one image, the identity pose and a depth estimate, here ZoeDepth~\cite{bhat2023zoedepth}. In this example, we register 33 views to the initial reconstruction. Depth estimates are only used for this step.}
   \label{fig:pose_refine_seed}
\end{figure*}

\noindent \textbf{Pose Refinement.}
Differently from the ACE~\cite{brachmann2023accelerated} protocol, the ground truth poses $\pose^{\text{GT}}_i$ are unknown during training.
Instead, we have $\pose^t_i$, estimates based on earlier iterations of the reconstruction.
Since these estimates can be inaccurate, we add the ability to refine poses during mapping.
We implement refinement using an MLP:
\begin{equation}
    \pose_i^t = f^t_\text{Pose}(\tilde{\pose}^t_i),
\end{equation}
where $\tilde{\pose}^t_i$ denotes the initial pose estimate at the start of a mapping iteration.
Inspired by \cite{Zhou2019Continuity}, the refinement MLP ingests $\tilde{\pose}^t_i$ as $3\times4=12$ values and predicts 12 additive offsets.
We orthonormalize rotations using Gram-Schmidt~\cite{bregier2021deepregression}.
We jointly optimize $f^t_\text{Pose}$ and  $f^t_\text{SCR}$ to minimize the reprojection error of Eq.~\ref{eq:repro_objective}.
We show the impact of pose refinement for one iteration in Fig.~\ref{fig:pose_refine_seed} (left).
We discard the refinement MLP after each mapping iteration.
Its purpose is to enable the scene model, $f^t_\text{SCR}$, to converge to a consistent scene representation.

With neither poses nor 2D-3D correspondences fixed in Eq.~\ref{eq:repro_objective}, the scene coordinate regressor could drift or degenerate.
As regularization, we optimize the pose refiner $f^t_\text{Pose}$ using AdamW~\cite{loshchilov2019decoupled} with weight decay, biasing the MLP to predict small updates relative to the initial estimates $\tilde{\pose}^t_i$.
This relies on the assumption that mapping images in $\mathcal{I}^t_\text{Reg}$ have been registered close to their true position.
If that assumption holds, the smoothness prior of the networks~\cite{ulyanov2018deep} encourages a multi-view consistent solution as shown in previous work \cite{brachmann2018dsacpp, brachmann2021dsacstar, brachmann2023accelerated}.

As an alternative to an MLP refiner, we could back-propagate directly to the input poses~\cite{yen2021inerf,wang2021nerf}.
However, when optimizing over thousands of views, the signal for each single pose becomes sparse.
Cameras are correlated via the scene representation.
If the optimization removes drift in the scene, multiple cameras need to move.
The MLP refiner models the correlation of cameras.

\noindent \textbf{Calibration Refinement.}
We do not assume information about precise calibration parameters, although often reported by devices.
We do assume that the principal point is in the center, that pixels are unskewed and square.
We do not model image distortion.
While these are reasonable assumptions for many data regimes, we cannot rely on the focal length to be given.
Thus, we refine the focal length starting from a heuristic: $K_i^t=f^t_\text{Calib}(K^\text{init}_i)$.
The refinement function $f^t_\text{Calib}$ entails a single learnable parameter $\alpha^t$ such that the focal length $f^t_i = f^\text{init} \cdot (1 + \alpha^t)$.
As before, superscript $t$ denotes the time step.
We optimize $\alpha^t$ using AdamW~\cite{loshchilov2019decoupled} with weight decay, biasing it towards a small relative scale factor.
Estimates of $\alpha$ are carried over across iterations.
We set $f^\text{init}$ to 70\% of the image diagonal and it is shared by all cameras in our experiments.

\subsection{Relocalization}

Given the scene model of iteration $t$, we attempt to register more images to determine the training set for the next mapping iteration, $\mathcal{I}^{t+1}_\text{Reg}$.
We pass all images in $\mathcal{I}$ to the scene coordinate regressor $f^t_\text{SCR}$ to gather 2D-3D correspondences, and solve for their poses using RANSAC and PnP:
\begin{equation}
    \tilde{\pose}^{t+1}_i, s^{t+1}_i = g\left(K^t_i, \left\{\left(\pos_{ij}, \crd^t_{ij}\right)\right\}\right).
\end{equation}
Here, we assume that the pose solver returns a confidence score $s^{t+1}_i$ alongside the pose itself that lets us decide whether the pose of the image has been estimated successfully.
We simply utilize the inlier count as score $s^{t+1}_i$ and apply a threshold to form the training set of the next mapping iteration: $\mathcal{I}^{t+1}_\text{Reg} = \left\{I_i | s^{t+1}_i > \tau_s \right\}$.
The relocalization process is depicted in \Cref{fig:system_ace0}, bottom.

\subsection{Initialization}
\label{sec:method:init}

We start the reconstruction with one image: $\mathcal{I}^{0}_\text{Reg}=\left\{I_\text{seed}\right\}$.
We set the seed pose $\pose^{0}_\text{seed}$ to identity, and we initialize the calibration $K^\text{init}_\text{seed}$ as explained above.

We cannot train a scene coordinate regression network using Eq.~\ref{eq:repro_objective} with a single image.
The reprojection error is ill-conditioned without multiple views constraining the depth.
Therefore, we optimize a different objective in the seed iteration, inspired by Map-free Relocalization~\cite{arnold2022map}.
Arnold~\etal argue that a single image and a depth estimate allow to relocalize query images, albeit with limited accuracy.
Our experiments show that such coarse relocalizations of a few images suffice as initialization for optimizing the reprojection error of Eq.~\ref{eq:repro_objective}.

Let $d_{ij}$ be a depth value predicted for pixel $j$ of image $i$. 
We derive a target scene coordinate by back-projection as $\hat{\crd}_{ij}=d_{ij} (K_\text{seed}^\text{init})^{-1} \pos_{ij}$.
We train an initial scene coordinate regression network $f^0_\text{SCR}$ by optimizing $\sum_{j \in I^0} \left\Vert \hat{\crd}_{ij} - \crd_{ij}\right\Vert$.
Fig.~\ref{fig:pose_refine_seed} (right) shows a scene coordinate point cloud learned from a depth estimate, and successful relocalization against it.

We found our pipeline to be robust \wrt selecting the seed image.
However, when a randomly selected image has little to no visual overlap with the remaining images, the whole reconstruction would fail.
Selecting an unfortunate seed image, \eg, at the very end of a long camera trajectory, can increase reconstruction times.
To decrease the probability of such incidents, we try 5 random seed images and choose the one with the highest relocalization rate across 1000 other mapping images. 
Since mapping seed images is fast, \emph{ca}.~1 min on average, this poses no significant computational burden.

\subsection{Implementation} 

We base our pipeline on the public code of the ACE relocalizer~\cite{brachmann2023accelerated}.
ACE uses a convolutional feature backbone, pre-trained on ScanNet~\cite{dai2017scannet}, that ingests images scaled to 480px height.
On top of the backbone is the mapping network, a 9-layer MLP with 512 channels, consuming 4MB of weights in 16-bit floating point precision.
This is our scene representation.

The ACE training process is reasonably fast, taking 5 minutes to train the scene network. 
Since we repeat training the scene network in multiple iterations, we expedite the process further to decrease our total reconstruction time.

\noindent \textbf{Adaptive Sampling of Training Patches.}
ACE trains the scene representation based on 8M patches sampled from the mapping images, a process that takes 1 minute.
This is excessive when having very few mapping images in the beginning of the reconstruction, thus we loop over the mapping images at most 10 times when sampling patches, or until 8M patches have been sampled.
Using this adaptive strategy, sampling patches for the seed reconstruction where we have 1 image only takes 2 seconds instead of 1 minute.

\noindent \textbf{Adaptive Stopping.}
ACE trains the scene model using a fixed one-cycle learning rate schedule~\cite{smith2019super} to a total of 25k parameter updates.
We make the training schedule adaptive since the network is likely to converge fast when trained on few images only.
We monitor the reprojection error of scene coordinates within a mini-batch.
If for 100 consecutive batches 70\% of reprojection errors are below an inlier threshold of 10px, we stop training early.
We approximate the one-cycle learning rate schedule in a linear fashion: we increase the learning rate in the first 1k iteration from $5\times10^{-4}$ to $3\times10^{-3}$ and when the early stopping criterion has been met, we decrease the learning rate to $5\times10^{-4}$ within 5k iterations.

We report more implementation details and hyper-parameters in the supplement. 
Our code is also publicly available to ensure reproducibility.

\section{Experiments}
\label{sec:experiments}

\begin{figure*}[t]
  \centering
   \includegraphics[width=1.0\linewidth]{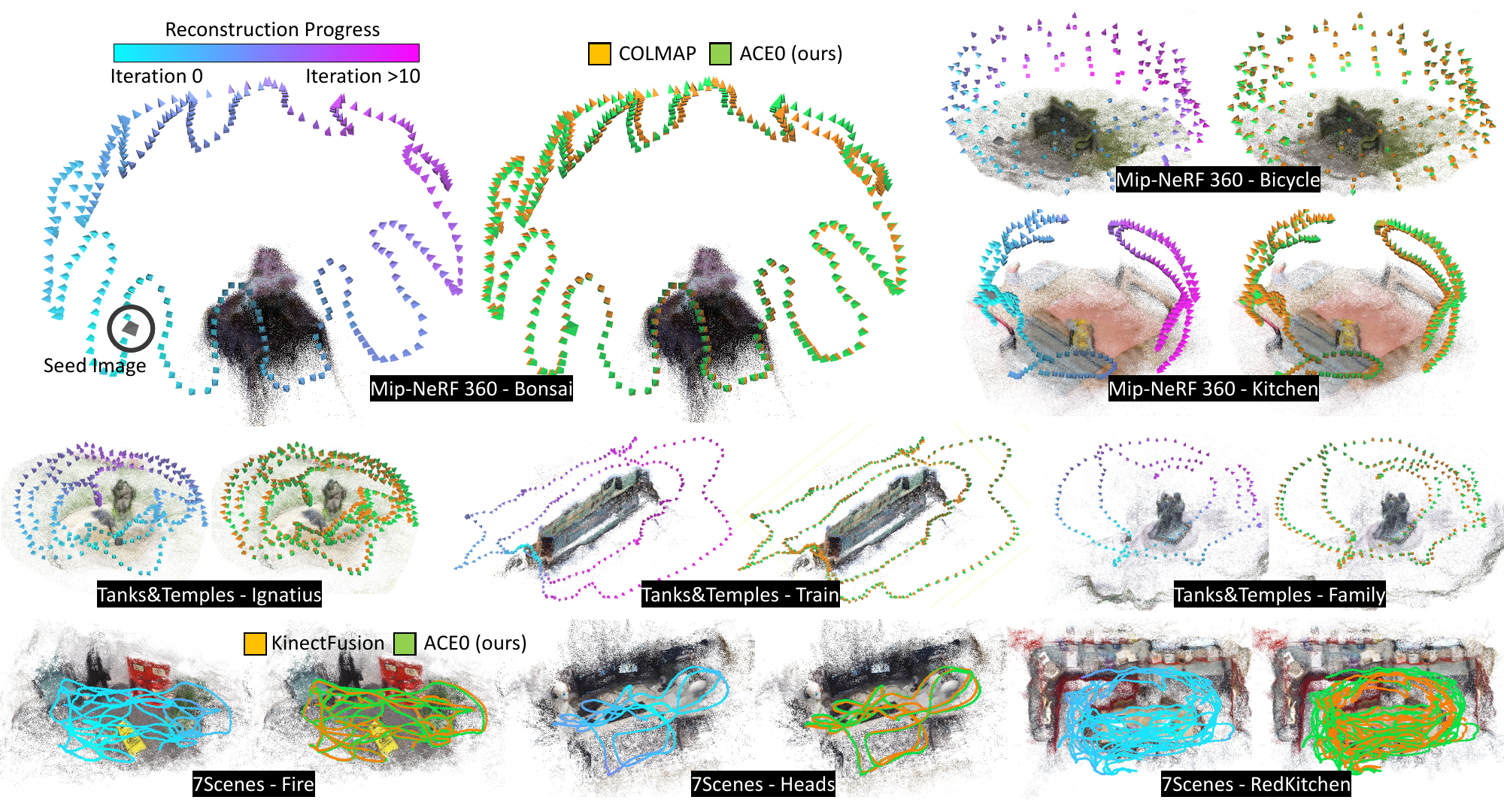}
   \caption{\textbf{Reconstructed Poses.} We show poses estimated by ACE0 for a selection of scenes. We color code the reconstruction iteration in which a particular view has been registered. We show the ACE0 point cloud as a representation of the scene. The seed image is shown as a gray frustum. We also compare our poses to poses estimated by COLMAP (Mip-NeRF 360, Tanks and Temples) and KinectFusion (7-Scenes). }
   \label{fig:exp:qualitative}
\end{figure*}

We refer to our SfM pipeline as ACE0 (ACE Zero) since it builds on top of the ACE relocalizer~\cite{brachmann2023accelerated} but adds the ability to train from scratch, without poses.
We demonstrate the effectiveness of our approach on three datasets and 31 scenes in total. 
For all experiments, we rely on ZoeDepth~\cite{bhat2023zoedepth} to initialize our reconstructions.
All timings reported for ACE0 are based on a single V100 GPU.

\noindent \textbf{Baselines.}
We consider the pose estimates of COLMAP with default parameters as our pseudo ground truth, obtained by extensive processing.
We also run COLMAP with parameter settings recommended for large image collections of 1k images and more~\cite{fastcolmap}. 
This variation achieves much faster processing (denoted \emph{COLMAP fast}).
We use a V100 GPU for COLMAP feature extraction and matching, and we specify the exact parameters of COLMAP in the supplement.
Furthermore, we show some results of RealityCapture \cite{reality-capture}, an efficient commercial feature-based SfM pipeline.

We compare to learning-based SfM approaches that are not restricted to few-frame scenarios, namely to NeRF-based BARF \cite{lin2021barf} and NoPe-NeRF~\cite{bian2023nope}.
We also compare to DUSt3R \cite{wang2024dust3r}, a non-NeRF learning-based SfM method.
While we focus on reconstructing unsorted image collections, the datasets we consider allow for sequential processing.
Hence, for context, we also show results of DROID-SLAM \cite{teed2021droid}, a neural SLAM approach. 
Unless specified otherwise, we report timings based on a single V100 GPU. 

\noindent \textbf{Benchmark.}
We show results on 7-Scenes~\cite{shotton2013scene}, a relocalization dataset, on Mip-NeRF 360~\cite{barron2022mipnerf360}, a view synthesis dataset and on Tanks and Temples~\cite{tanksandtemples}, a reconstruction dataset.
Comparing poses on these datasets is problematic, as our pseudo ground truth is estimated rather than measured.
For example, an approach might be more accurate than COLMAP on individual scenes. 
Computing pose errors \wrt COLMAP would result in incorrect conclusions.
Therefore, we gauge the pose quality in a self-supervised way, using novel view synthesis \cite{waechter2017rephotography}.

We let each method estimate the poses of all images of a scene. 
For evaluation, we split the images into training and test sets.
We train a Nerfacto~\cite{nerfstudio} model on the training set, and synthesize views for the test poses.
We compare the synthesized images to the test images and report the difference as peak signal-to-noise ratio (PSNR).
To ensure a fair comparison to NeRF-based competitors, we use these methods in the same way as the other SfM methods:
we run them to pose all images, and train a Nerfacto model on top of their poses.
This is to ensure that we solely compare the pose quality across methods, and not different capabilities in view synthesis.
The quality of poses affects the PSNR numbers in two ways:
Good training poses let the NeRF model fit a consistent scene representation.
Good testing poses make sure that the synthesized images are aligned with the original image.
We explain further details in the supplement which also includes additional, perceptual metrics.
In spirit, our evaluation is similar to the Tanks and Temples benchmark which evaluates a derived scene mesh rather than camera poses. 
However, our evaluation can be applied to arbitrary datasets as it does not need ground truth geometry.

\subsection{7-Scenes}
\label{sec:experiments:sevenscenes}

\begin{table*}[t]
\bgroup
    \centering%
    \footnotesize%
    \adjustbox{valign=t,width=\linewidth}{%
        \begin{tabular}{cc||cc||cccccccccccc||ccc||cc}
             & & & & & & & & & & & & & & & & & & & & \\ 
            \toprule
            \multicolumn{2}{c||}{~} &%
            \multicolumn{2}{c||}{~Pseudo Ground Truth~} &%
            \multicolumn{12}{c||}{~All Frames~} &%
            \multicolumn{3}{c||}{~200 Frames~} &%
            \multicolumn{2}{c}{~50 Frames~} \\
            \hhline{~~-------------------}
             ~ &%
            \hspace{-6pt}\parbox[t]{0mm}{\multirow{2}{*}{\rotatebox[origin=c]{90}{Frames\hspace{-12pt}}}} &%
            ~Kinect~ &%
            ~COLMAP~ &%
            \multicolumn{3}{c}{~COLMAP~} &%
            \multicolumn{3}{c}{~DROID-SLAM$^\dagger$~} &%
            \multicolumn{3}{c}{~ACE0~} &%
            \multicolumn{3}{c||}{~KF+ACE0~} &%
            ~BARF~ &%
            ~NoPE-NeRF$^\dagger$~ &%
            ~ACE0~ &%
            ~DUSt3R~ &%
            ~ACE0~ \\
             &%
             ~ &%
            Fusion &%
            (default) &%
            \multicolumn{3}{c}{(fast)} &%
            \multicolumn{3}{c}{\cite{teed2021droid}} &%
            \multicolumn{3}{c}{(ours)} &%
            \multicolumn{3}{c||}{(ours)} &%
            \cite{lin2021barf} &%
            \cite{bian2023nope} &%
            (ours) &%
            \cite{wang2024dust3r} &%
            (Ours) \\
            \midrule 
             Chess & 6k & 19.6 & 23.6 & \multicolumn{3}{c}{\begin{tabular}{ccc} & ~{\cellcolor{TableLightGreen}23.5}~ & \end{tabular}} &  \multicolumn{3}{c}{\begin{tabular}{ccc} & ~{\cellcolor{TableRed}19.3}~ & \end{tabular}} &  \multicolumn{3}{c}{\begin{tabular}{ccc} & ~{\cellcolor{TableLightGreen}23.3}~ & \end{tabular}} &  \multicolumn{3}{c||}{\begin{tabular}{ccc} & ~{\cellcolor{TableYellow}23.0}~ & \end{tabular}} &  12.8 & 12.6 & \textbf{22.7} & 18.9 &\textbf{19.2} \\
             & & & & & & & & & & & & & & & & & & & &\\[-10pt]
 Fire & 4k & 19.2 & 22.6 & \multicolumn{3}{c}{\begin{tabular}{ccc} & ~{\cellcolor{TableLightGreen}22.6}~ & \end{tabular}} &  \multicolumn{3}{c}{\begin{tabular}{ccc} & ~{\cellcolor{TableRed}13.0}~ & \end{tabular}} &  \multicolumn{3}{c}{\begin{tabular}{ccc} & ~{\cellcolor{TableLightGreen}22.3}~ & \end{tabular}} &  \multicolumn{3}{c||}{\begin{tabular}{ccc} & ~{\cellcolor{TableLightGreen}22.3}~ & \end{tabular}} &  12.7 & 11.8 & \textbf{22.1} & 18.8 &\textbf{19.5} \\
             & & & & & & & & & & & & & & & & & & & &\\[-10pt]
 Heads & 2k & 17.0 & 18.8 & \multicolumn{3}{c}{\begin{tabular}{ccc} & ~{\cellcolor{TableLightGreen}18.9}~ & \end{tabular}} &  \multicolumn{3}{c}{\begin{tabular}{ccc} & ~{\cellcolor{TableRed}17.6}~ & \end{tabular}} &  \multicolumn{3}{c}{\begin{tabular}{ccc} & ~{\cellcolor{TableLightGreen}18.8}~ & \end{tabular}} &  \multicolumn{3}{c||}{\begin{tabular}{ccc} & ~{\cellcolor{TableLightGreen}19.1}~ & \end{tabular}} &  10.7 & 11.8 & \textbf{19.9} & 18.4 &\textbf{21.3} \\
             & & & & & & & & & & & & & & & & & & & &\\[-10pt]
 Office & 10k & 18.9 & 21.4 & \multicolumn{3}{c}{\begin{tabular}{ccc} & ~{\cellcolor{TableLightGreen}21.6}~ & \end{tabular}} &  \multicolumn{3}{c}{\begin{tabular}{ccc} & {\cellcolor{TableRed}failed} & \end{tabular}} &  \multicolumn{3}{c}{\begin{tabular}{ccc} & ~{\cellcolor{TableLightGreen}21.1}~ & \end{tabular}} &  \multicolumn{3}{c||}{\begin{tabular}{ccc} & ~{\cellcolor{TableLightGreen}21.5}~ & \end{tabular}} &  11.9 & 10.9 & \textbf{19.8} & 12.5 &\textbf{13.7} \\
             & & & & & & & & & & & & & & & & & & & &\\[-10pt]
 Pumpkin & 6k & 19.9 & 24.1 & \multicolumn{3}{c}{\begin{tabular}{ccc} & ~{\cellcolor{TableLightGreen}23.8}~ & \end{tabular}} &  \multicolumn{3}{c}{\begin{tabular}{ccc} & ~{\cellcolor{TableRed}18.3}~ & \end{tabular}} &  \multicolumn{3}{c}{\begin{tabular}{ccc} & ~{\cellcolor{TableLightGreen}24.1}~ & \end{tabular}} &  \multicolumn{3}{c||}{\begin{tabular}{ccc} & ~{\cellcolor{TableLightGreen}23.8}~ & \end{tabular}} &  19.6 & 14.2 & \textbf{24.7} & 21.7 &\textbf{22.3} \\
             & & & & & & & & & & & & & & & & & & & &\\[-10pt]
 RedKitchen & 12k & 17.6 & 21.4 & \multicolumn{3}{c}{\begin{tabular}{ccc} & ~{\cellcolor{TableLightGreen}21.4}~ & \end{tabular}} &  \multicolumn{3}{c}{\begin{tabular}{ccc} & ~{\cellcolor{TableRed}10.9}~ & \end{tabular}} &  \multicolumn{3}{c}{\begin{tabular}{ccc} & ~{\cellcolor{TableYellow}20.8}~ & \end{tabular}} &  \multicolumn{3}{c||}{\begin{tabular}{ccc} & ~{\cellcolor{TableLightGreen}20.9}~ & \end{tabular}} &  11.6 & 11.2 & \textbf{18.9} & \textbf{13.8} &13.7 \\
             & & & & & & & & & & & & & & & & & & & &\\[-10pt]
 Stairs & 3k & 19.0 & 16.7 & \multicolumn{3}{c}{\begin{tabular}{ccc} & ~{\cellcolor{TableDarkGreen}21.0}~ & \end{tabular}} &  \multicolumn{3}{c}{\begin{tabular}{ccc} & ~{\cellcolor{TableRed}13.0}~ & \end{tabular}} &  \multicolumn{3}{c}{\begin{tabular}{ccc} & ~{\cellcolor{TableDarkGreen}17.7}~ & \end{tabular}} &  \multicolumn{3}{c||}{\begin{tabular}{ccc} & ~{\cellcolor{TableDarkGreen}19.9}~ & \end{tabular}} &  15.8 & 15.9 & \textbf{18.8} & 15.3 &\textbf{15.4} \\
             & & & & & & & & & & & & & & & & & & & &\\[-10pt]
\hhline{~--------------------}
             & & & & & & & & & & & & & & & & & & & &\\[-10pt]
             Average &      & 18.7 & 21.2 & \multicolumn{3}{c}{\begin{tabular}{ccc} & ~{\cellcolor{TableDarkGreen}21.8}~ & \end{tabular}} &  \multicolumn{3}{c}{\begin{tabular}{ccc} & {\cellcolor{TableRed}{\color{TableRed}0}N/A} & \end{tabular}} &  \multicolumn{3}{c}{\begin{tabular}{ccc} & ~{\cellcolor{TableLightGreen}21.2}~ & \end{tabular}} &  \multicolumn{3}{c||}{\begin{tabular}{ccc} & ~{\cellcolor{TableLightGreen}21.5}~ & \end{tabular}} &  13.6 & 12.6 & \textbf{21.0} & 17.1 &\textbf{17.9} \\
             & & & & & & & & & & & & & & & & & & & &\\[-10pt]
\hline
             Avg. Time &      & realtime & 38h & \multicolumn{3}{c}{\begin{tabular}{ccc} & ~13h~ & \end{tabular}} &  \multicolumn{3}{c}{\begin{tabular}{ccc} & ~18min~ & \end{tabular}} &  \multicolumn{3}{c}{\begin{tabular}{ccc} & ~1h~ & \end{tabular}} &  \multicolumn{3}{c||}{\begin{tabular}{ccc} & ~7min~ & \end{tabular}} &  8.5h & 47h & \textbf{27min} & \textbf{4min*} &16min \\
             & & & & & & & & & & & & & & & & & & & &\\[-10pt]
\bottomrule
      \end{tabular}
    }
    \caption{
        \textbf{7-Scenes.} 
        We show the pose accuracy via view synthesis with Nerfacto~\cite{nerfstudio} as PSNR in dB, and the reconstruction time.
        Results for \emph{All Frames} are color coded \wrt similarity to the COLMAP pseudo ground truth:
        \TCBTableDarkGreen{$>0.5$ dB better}\TCBTableLightGreen{within $\pm 0.5$ dB}\TCBTableYellow{$>0.5$ dB worse}\TCBTableRed{$>$1 dB worse}.
        For some competitors, we had to sub-sample the images due to their computational complexity (right side). $^{\dagger}$Method needs sequential inputs. $^{*}$Results on more powerful hardware.
    }%
    \label{tab:sevenscenes}%
\egroup

\end{table*}

The 7-Scenes dataset~\cite{shotton2013scene} consists of seven indoor scenes, scanned with a Kinect v1 camera.
Multiple, disconnected scans are provided for each scene to a total of 2k-12k images.
For each method, we assume a shared focal length across scans and initialize with the default calibration of a Kinect v1. 
The dataset comes with pseudo ground truth camera poses estimated by KinectFusion~\cite{izadi2011kinectfusion,newcombe2011kinectfusion}, a depth-based SLAM system.
Individual scans were registered but not bundle-adjusted~\cite{brachmann2021limits}.
Inspired by \cite{brachmann2021limits}, we recompute alternative, bundle-adjusted pseudo ground truth by running COLMAP with default parameters.

\noindent \textbf{Discussion.}
We show results in \Cref{tab:sevenscenes}.
Of both pseudo ground truth versions, KinectFusion achieves lower PSNR numbers than COLMAP, presumably due to the lack of global optimization.
COLMAP with \emph{fast} parameters shows PSNR numbers similar to COLMAP with \emph{default} parameters, on average.
Both versions of running COLMAP take considerable time to reconstruct each scene.
We note that COLMAP has been optimised for quality, rather than speed. 
Not all acceleration strategies from the feature-based SfM literature have been implemented in COLMAP, so presumably comparable quality can be obtained faster. 
DROID-SLAM \cite{teed2021droid} does not perform well on 7-Scenes and partially fails altogether, presumably due to the jumps between individual scans of each scene.

Our approach, ACE0, achieves a pose quality comparable to the COLMAP pseudo ground truth while reconstructing each scene in $\sim$1 hour despite the large number of images.
We show qualitative examples in \Cref{fig:exp:qualitative} and in the supplement.
We also demonstrate that ACE0 can swiftly optimize an initial set of approximate poses.
When starting from KinectFusion poses, ACE0 increases PSNR significantly in less than 10 minutes per scene, see ``KF+ACE0'' in \Cref{tab:sevenscenes}.
In the supplement, we include a parameter study on 7-Scenes to show that ACE0 is robust to the choice of depth estimator.
We also show the positive impact of pose refinement on the reconstruction quality as well as the reconstruction speedup due to our early stopping schedule.

For our learning-based competitors, we sub-sampled images due to their computational constraints, see right side of Table~\ref{tab:sevenscenes}.
Even using only 200 images, NoPe-NeRF \cite{bian2023nope} takes 2 days to fit a model and estimate poses. 
Despite these long processing times, we observe poor pose quality  of BARF and NoPe-NeRF in terms of PSNR.
BARF \cite{lin2021barf} requires pose initialisation.
We provide identity poses since the scenes of 7Scenes are roughly forward-facing.
Still, the camera motions are too complex for BARF to handle.
NoPe-NeRF does not require pose initialisation but needs roughly sequential images, which we did provide.
NoPe-NeRF relies upon successive images having similar poses to perform well, and struggles with the large jumps between the subsampled images.

For the comparison with DUSt3R \cite{wang2024dust3r}, we had to subsample the sequences further, as we were only able to run it with 50 images at most, even when using an A100 GPU with 40GB memory.
DUSt3R achieves reasonable PSNR numbers but consistently lower than ACE0. 

\noindent \textbf{Relocalization.}
ACE0 is a learning-based SfM tool but it is also a self-supervised visual relocaliser. 
In \subref{tab:sevenscenes_reloc}, we compare it to the supervised relocalizer ACE \cite{brachmann2023accelerated}.
Using the scale-metric pseudo ground truth of \cite{brachmann2021limits}, we train ACE with COLMAP mapping poses, and evaluate it against COLMAP query poses. 
Unsurprisingly, ACE achieves almost perfect relocalization under the usual \mbox{5cm, 5$^\circ$} error threshold.
Interestingly, ACE0 achieves almost identical results when mapping the scene self-supervised, and evaluating the relocalized query poses against the COLMAP pseudo ground truth. 
For context, when training ACE with KinectFusion mapping poses and evaluating against COLMAP pseudo ground truth, results are far worse. 
This signifies that ACE0 mapping poses are very similar to the COLMAP mapping poses, and less similar to KinectFusion mapping poses.
We give more details about this experiment in the supplement.

\begin{table*}[t]
    \bgroup%
    \footnotesize%
    \adjustbox{valign=t,width=\linewidth}{%
        \sisetup{detect-all=true,detect-weight=true}
        \begin{tabular}{
        @{}
        l
        S[table-format=3.1, table-align-text-after=false, table-space-text-post = {\percent}]
        S[table-format=3.1, table-align-text-after=false, table-space-text-post = {\percent}]
        S[table-format=3.1, table-align-text-after=false, table-space-text-post = {\percent}]
        c
        lc||cccc
        @{}
        }%
            \hhline{----~------}
            \hhline{----~------}
            %
            %
            %
            & \multicolumn{1}{@{}c@{}}{~ACE \cite{brachmann2023accelerated}~}
            & \multicolumn{1}{@{}c@{}}{~ACE \cite{brachmann2023accelerated}~}
            & \multicolumn{1}{@{}c@{}}{~ACE0 (ours)~}%
            & \parbox{0.15cm}{~} 
            %
            %
            & 
            & \multicolumn{1}{c||}{~Pseudo GT~}%
            & ~DROID-SLAM$^\dagger$~
            & ~BARF~
            & ~NoPe-NeRF$^\dagger$~
            &\multicolumn{1}{c}{~ACE0~}%
            \\%
            %
            %
              \multicolumn{1}{@{}c@{}}{Supervision~}%
            & \multicolumn{1}{@{}c@{}}{~KinectFusion~}%
            & \multicolumn{1}{@{}c@{}}{~COLMAP~}%
            & \multicolumn{1}{@{}c@{}}{~\textit{self-supervised}~}%
            & 
            %
            %
            & 
            &\multicolumn{1}{c||}{(COLMAP)}%
            & \cite{teed2021droid}%
            & \cite{lin2021barf}%
            & \cite{bian2023nope}%
            &\multicolumn{1}{c}{(ours)}%
            \\%
            %
            %
            \hhline{----~------}
            Chess & ~~\underline{\SI{96.0}{\percent}} & \bfseries{\SI{100.0}{\percent}} & \bfseries{\SI{100.0}{\percent}} & 
            & 
            Bicycle	& 21.5 & 10.9 & 11.9 & 12.2 & \textbf{18.7} \\%
            Fire       & ~~\SI{98.4}{\percent} & ~~\bfseries{\SI{99.5}{\percent}} & ~\underline{\SI{98.8}{\percent}} &
            & 
            Bonsai	& 27.6	& 10.9 & 12.5 & 14.8 & \textbf{25.8} \\%
            Heads      & \bfseries \SI{100.0}{\percent} & \bfseries{\SI{100.0}{\percent}} & \bfseries \SI{100.0}{\percent} &
            & 
            Counter & 25.5	& 12.9 & 11.9 & 11.6 & \textbf{24.5} \\%
            Office     & ~~\SI{36.9}{\percent} & \bfseries{\SI{100.0}{\percent}} & ~\underline{\SI{99.1}{\percent}} &
            & 
            Garden	& 26.3	& 16.7 & 13.3 & 13.8 & \textbf{25.0} \\%
            Pumpkin    & ~~\SI{47.3}{\percent} & \bfseries{\SI{100.0}{\percent}} & ~\underline{\SI{99.9}{\percent}} &
            & 
            Kitchen & 27.4 & 13.9 & 13.3 & 14.4 & \textbf{26.1} \\%
            Redkitchen & ~~\SI{47.8}{\percent} & ~~\bfseries{\SI{98.9}{\percent}} & ~\underline{\SI{98.1}{\percent}} &
            & 
            Room	& 28.0	& 11.3 & 11.9 & 14.3 & \textbf{19.8} \\%
            Stairs     & ~~\underline{\SI{74.1}{\percent}} & ~~\bfseries{\SI{85.0}{\percent}} & ~\SI{61.0}{\percent} &
            & 
            Stump	& 16.8 & 13.9 & 15.0 & 13.7 & \textbf{20.5} \\%
            \hhline{----~------}
            Average    & ~~\SI{71.5}{\percent} & ~~\bfseries{\SI{97.6}{\percent}} & ~\underline{\SI{93.8}{\percent}} & 
            & 
            Average   & 24.7 & 12.9 & 12.8 &	13.5 &	\textbf{22.9} \\%
        \end{tabular}%
    }
    \\[-2pt]
    \begin{subcaptionblock}[T]{.41\linewidth}
        \caption{%
            \textbf{Relocalization on 7-Scenes.} \% poses below \mbox{5cm, 5$^\circ$} error, computed \wrt COLMAP pseudo GT.%
        }\label{tab:sevenscenes_reloc}
    \end{subcaptionblock}
    \hfill
    \begin{subcaptionblock}[T]{.55\linewidth}
        \caption{%
          \textbf{Mip-NeRF 360.} Pose quality in PSNR, higher is better. Best in \textbf{bold}. $^\dagger$Method needs sequential inputs.%
        }\label{tab:mip}
    \end{subcaptionblock}
\egroup%
\end{table*}

\subsection{Mip-NeRF 360}
\label{sec:experiments:mip}

The Mip-NeRF 360 dataset~\cite{barron2022mipnerf360} consists of seven small-scale scenes, both indoor and outdoor.
The dataset was reconstructed with COLMAP and comes with intrinsics (which we ignore), camera poses and undistorted (pinhole camera) images.
For each method, we assume a shared focal length per scene.

\noindent \textbf{Discussion.}
We present PSNR results in \subref{tab:mip}.
NoPe-NeRF does not perform well on this dataset, despite processing each scene for 2 days.
The differences between sequential images are too large.
DROID-SLAM fails for the same reason.
BARF performs poorly because the identity pose is not a good initialisation for most scenes.
In contrast, ACE0 reconstructs the dataset successfully.
While it achieves slightly lower PSNR than COLMAP, its pose estimates are similar, \cf \Cref{fig:exp:qualitative}.
The supp.~shows that synthesized images based on ACE0 are visually close to those of COLMAP while our learning-based competitors are far off.

\subsection{Tanks and Temples}
\label{sec:experiments:t2}

\begin{table*}[t]%
    \centering%
    \setlength{\tabcolsep}{1pt}%
    \footnotesize%
    \adjustbox{valign=t,width=\linewidth}{%
} \\
        \bottomrule
    \end{tabular}
    }
    \caption{\textbf{Tanks and Temples.} We show the pose accuracy via view synthesis with Nerfacto~\cite{nerfstudio} as PSNR in dB, and the reconstruction time. We color code results compared to COLMAP, \emph{default} and \emph{fast}, respectively: \TCBTableDarkGreen{$>0.5$ dB better}\TCBTableLightGreen{within $\pm 0.5$ dB}\TCBTableYellow{$>0.5$ dB worse}\TCBTableRed{$>$1 dB worse}. $^{\dagger}$Method needs sequential inputs.}%
    \label{tab:t2}%
\end{table*}

The Tanks and Temples dataset~\cite{tanksandtemples} contains 21 diverse scenes, both indoors and outdoors, and with varying spatial extent.
We remove two scenes (Panther, M60) with military associations. 
We also remove two scenes (Courthouse, Museum) where the COLMAP baseline did not finish the reconstruction after 5 days of processing or ran out of memory.
For the latter two scenes, we provide ACE0 results in the supplement.
The dataset provides 150-500 images per scene but also the original videos as a source for more frames.
Thus, we consider each scene in two versions: Using 150-500 images and using 4k-22k frames.
For all methods, we assume a pinhole camera model with shared focal length across images, and images to be unordered.
We found none of the learning-based SfM competitors to be applicable to this dataset.
NoPe-NeRF would run multiple days per scene, even when considering only a few hundred images.
DUSt3R would run out of memory. 
BARF needs reasonable pose initializations which are not available.

\noindent \textbf{Discussion.}
We show PSNR numbers of RealityCapture, DROID-SLAM and ACE0 in \Cref{tab:t2}, color-coded by similarity to the COLMAP pseudo GT.
ACE0 achieves reasonable results when reconstructing scenes from a few hundred images (\Cref{tab:t2}, left).
ACE0 generally predicts plausible poses (\cf, \Cref{fig:exp:qualitative}), even if PSNR numbers are sometimes lower than those of COLMAP.
RealityCapture performs similar to ACE0 while DROID-SLAM struggles on the sparse images.

Next, we consider more than 1k images per scene (right side of \Cref{tab:t2}).
Here, we run COLMAP with parameters tuned for large images collections (\emph{fast)} due to the extremely large image sets.
ACE0 offers a reconstruction quality comparable to COLMAP on average while also being fast.
We run RealityCapture on some of the scenes but it produces fractured reconstructions for these large images sets, leading to low PSNR numbers.
DROID-SLAM still struggles on many scenes despite having access to sequential images that are temporally close.

In the two rightmost columns, we initialise with a sparse COLMAP reconstruction from 150-500 images, and extend and refine it using all available frames.
Firstly, using a feature-based baseline, we register the full set of frames using the relocalization mode of COLMAP, followed by a final round of bundle adjustment.
Secondly, we run ACE0 initialised with the poses of the sparse COLMAP reconstruction. 
Both variants are considerably faster than running COLMAP from scratch on the full set of frames.
ACE0 is able to register and refine all additional frames in 1-2 hours, on average.
Again, we find the pose quality of ACE0 comparable to the feature-based alternative.

\section{Conclusion and Future Work}
\label{sec:conclusion}

We have presented scene coordinate reconstruction, a new approach to learning-based SfM.
We learn an implicit, neural scene representation from a set of unposed images.
Our method, ACE0, is able to reconstruct a wide variety of scenes.
In many cases, the accuracy of estimated poses is close to that of COLMAP and synthesized images visually similar.
Unlike previous learning-based SfM methods, ACE0 can be applied to multiple thousand unsorted images, without pose priors, and reconstructs them within a few hours.

\noindent \textbf{Limitations.}
We show some failure cases in the supplement.
Scene coordinate regression struggles with repetitive structures since the network is not able to make multi-modal predictions for visually ambiguous inputs.
Scene coordinate regression also struggles with representing large areas.
The common solution is to use network ensembles based on pre-clustering of the scene~\cite{brachmann2019expert,brachmann2023accelerated} which is difficult in a reconstruction setting.
While scene coordinate regression generalises quite well, it has difficulties to bridge extreme view point or lighting changes, such as day versus night.
In our experiments, we assumed a simple pinhole camera model with shared intrinsics across images.
To the best of our knowledge, scene coordinate regression has not been coupled with image distortion, thus far. 

\clearpage
\appendix
{
    \centering
    \Large
    \vspace{0.5em}Supplementary Material \\
    \vspace{1.0em}
}

\section{Implementation}

\paragraph{Reconstruction Loop.}
We start the reconstruction as explained in \Cref{sec:method:init}. 
We repeat neural mapping and registration of new views until all views have been registered, or less than 1\% of additional views are registered compared to the last relocalization stage.
In each relocalization iteration, we re-estimate the poses of all views, even if they have been registered before.
We also tried a variant that skips estimating poses that have been registered before.
But this version achieved slightly lower quality while not giving a significant speed advantage.
Re-estimating all poses gives the pipeline the ability to reduce inaccuracies and to correct previous outlier estimates.
We consider an image to be registered successfully, if it has at least 500 inlier correspondences with an inlier threshold of 10px.
At the end of the reconstruction, we apply a more conservative threshold of 1000 inliers to make sure only high quality poses are kept.
For each neural mapping iteration, we load the network weights of the scene coordinate regression network of the previous mapping iteration and refine further.
We conclude the reconstruction loop with a final refit of the model, using the standard ACE training parameters.
In this last iteration, we start with a randomly initialised network, rather than loading previous weights.

\paragraph{Neural Mapping.}
The ACE backbone predicts 512-dimensional features densely for the input image, but sub-sampled by a factor of 8.
Training images are augmented using random scaling and rotation as well as brightness and contrast changes as in ACE \cite{brachmann2023accelerated}.
When filling the training buffer, we randomly choose 1024 features per training image, and we sample features from each training image at most 10 times, \ie for 10 different augmentations.
During training, we use a batch size of 5120 features. 
Our pose refinement MLP has 6 layers with 128 channels and a skip connection between layer 1 and 3. 
We optimize the pose refinement MLP, and the relative scaling factor for the focal length, with separate AdamW \cite{loshchilov2019decoupled} instances and a learning rate of $10^{-3}$.
In the seed iteration, we optimize neither the seed pose (which is identity) nor the initial calibration parameters.

ACE uses a soft-clamped version of the reprojection error in \mbox{Eq.~\ref{eq:repro_objective}} with a soft threshold that reduces from 50px to 1px throughout training.
We use the same, soft-clamped loss but keep the threshold fixed at 50px as we saw no benefit when using the loss schedule but it complicates early stopping.
During the seed iteration, we switch from optimizing the Euclidean distance to optimizing the reprojection error if the reprojection error of a scene coordinate is lower than 10px. 
This hybrid objective is inspired by DSAC* \cite{brachmann2021dsacstar}.

When starting neural mapping from a set of initial poses, \eg from KinectFusion, a sparse COLMAP reconstruction or in the final model refit of the reconstruction loop, we keep the pose refinement MLP frozen for the first 5000 iterations.
The refinement MLP cannot predict sensible updates for an uninitialised scene representation, and the reconstruction would become unstable.
Note that this standby time is not required in all but the last training iteration, as we initialise the scene coordinate regressor using the weights of the last iteration, so the pose refiner can predict sensible updates right away.

\paragraph{Relocalization.}
For registering images to the scene reconstruction, we utilize the pose optimization of ACE \cite{brachmann2023accelerated}.
It consists of a PnP solver within a RANSAC loop.
ACE samples 64 pose hypotheses per image.
If sampling a hypothesis fails, ACE tries again, up to 1M times.
After sampling, ACE ranks hypotheses according to inlier counts using a reprojection error threshold of 10px.
The best hypothesis is refined by iteratively resolving PnP on the growing inlier set.

Usually, this procedure is fast, taking approximately 30ms per image.
However, when the predicted image-to-scene correspondences are bad, sampling hypothesis can take a long time due to the large number of re-tries.
For ACE0, this will happen often, namely when we try to register images that are not yet covered by the reconstruction.
Thus, we decrease the number of re-tries to from 1M to 16.
When registering images throughout the reconstruction, we also decrease the number of hypotheses drawn from 64 to 32.
This yields registration times of 20ms per image, irrespective of whether the registration succeeds or fails.

\paragraph{Runtime Complexity.}
The reconstruction time depends on the number of images but also on the spatial distribution of cameras.
In the worst case, complexity is O($n^2$).
\Eg a long camera trajectory without intersections or loops would require proportional to $n$ ACE0 iterations, with $n$ relocalizations in each step, giving O($n^2$).
The best case complexity is $\Omega(n)$, \eg for a forward facing scene where all images can be registered in 1-2 ACE0 iterations independent of the number of images.
In practise, we observe attractive run times of ACE0.

\section{Benchmark}
\label{sec:app:benchmark}

We benchmark the quality of each set of poses by training a NeRF model on the train views and evaluating on the test views. We use Nerfstudio's Nerfacto model for this~\cite{nerfstudio}. Where datasets already include a train/test split -- as in 7-Scenes -- we use that split. Where a dataset does not already include a train/test split, we take every eighth view as a test view, which is a common choice for many NeRF datasets.

Many scenes in this work have large test sets of several thousand images or more. This can lead to long evaluation times due to the cost of rendering thousands of test views with NeRF -- much longer than the time to fit the NeRF itself. For this reason we subsample very large test sets so that they contain no more than 2k views. To ensure that this reduced test set is representative, we do this by subsampling evenly throughout the full-size test set in a deterministic manner. The code to compute the resulting splits will be included in our code release.

We use Nerfacto's default normalization of poses to fit into a unit cube, and enable scene contraction so that distant elements of the scene can be modelled more accurately. We train each scene for 30k iterations. For NeRF training and evaluation we downscale the resolution so that the longest side of an image is not greater than 640 pixels, as NeRFs often perform poorly with high-resolution training images.

Where a pose for a test view is not provided by a given method, we use the identity pose for that view. This results in a very low PSNR for that image, penalising the method for its failure to provide a pose. This means that we evaluate all methods on the full set of test views, even where they do not provide poses for all such views. 
If a reconstruction results in multiple disconnected components, we use the largest component in our benchmark and consider the other frames as missing.
This happens rarely for COLMAP, but more often for RealityCapture (\emph{c.f.}~Sec.~\ref{sec:app:baselines}).

ACE0 pose estimates are generally available for all images, but we consider an image registered if its confidence exceeds the threshold of 1000 inliers.
However, when evaluating test images we always take the estimated pose instead of the identity pose as basis for the evaluation, even if the confidence is lower than 1000.
Note that the question of whether a frame is considered ``registered'' or not is still important when fitting the NeRF, as such images are excluded from NeRF training.
See Sec.~\ref{sec:app:results} for further information about registration rates for different methods.

\paragraph{Relocalization.}
In the following, we give more information about the relocalization experiment of \Cref{sec:experiments:sevenscenes} and \subref{tab:sevenscenes_reloc}.
We conducted the experiment on the 7-Scenes dataset adhering to its training / test split. 
Our experiments utilize the pseudo ground truth (pGT) poses of KinectFusion that come with the dataset, as well as the COLMAP pGT of \cite{brachmann2021limits}.
The authors of \cite{brachmann2021limits} obtained the COLMAP pGT by reconstructing the training images of each scene with COLMAP from scratch.
To make the pGT scale-metric, they aligned the COLMAP reconstruction with the KinectFusion poses.
To obtain poses for the test images of each scene as well, they registered the test images to the training reconstruction.
Lastly, they ran a final round of bundle adjustment to refine the test poses while keeping the training poses fixed.

In our experiment, we train the ACE relocalizer on the training images of each scene.
Firstly, we train ACE using the KinectFusion poses that come with the dataset.
Secondly, we train ACE using the COLMAP pGT of \cite{brachmann2021limits}.
We evaluate both version of the relocalizer using the COLMAP pGT.
As expected, ACE trained with KinectFusion poses scores lower than ACE trained with COLMAP pGT, when evaluating against COLMAP pGT.
This confirms the finding of \cite{brachmann2021limits} that KinectFusion and COLMAP produce different pose estimates.

Next, we run ACE0 on the training set of each scene to reconstruct it and estimate the training poses. 
One output of ACE0 is an ACE relocalizer, trained from scratch without poses, in a self-supervised fashion.
We apply the ACE0 relocalizer to the test images to estimate their poses - without any further model refinement or joint optimization.
The ACE0 reconstruction is scale-less, or only roughly scaled based on the ZoeDepth estimate of the seed image.
To enable the scale-metric evaluation of \subref{tab:sevenscenes_reloc}, we align the ACE0 pose estimates to the test pseudo ground truth by fitting a similarity transform. 
Specifically, we run RANSAC \cite{ransac} with a Kabsch \cite{kabsch1976solution} solver on the estimated and the ground truth camera positions with 1000 iterations.
Due to the optimization of the similarity transform, we might underestimate the true relocalization error.
For a fair comparison, we perform the exact same evaluation procedure for our ACE baselines and fit a similarity transform although they have already been trained with scale-metric poses.
The effect of fitting the similarity transform is small.
ACE, trained on COLMAP pGT, achieves 97.1\% average relocalization accuracy when we fit the similarity transform, and 97.6\% average accuracy without fitting the transform.
We compute accuracy as the percentage of test images where the pose error is smaller than 5cm and 5$^\circ$ compared to ground truth.
The ACE0 relocalizer achieves 93.8\% average accuracy when evaluated against the COLMAP pGT.
The accuracy differences are smaller than 1\% for all scenes except Stairs.
This experiment shows that ACE0 is a viable self-supervised relocalizer, and that ACE0 poses are very close to COLMAP poses up to a similarity transform - much closer than \eg KinectFusion poses are to COLMAP poses.

\begin{figure*}[h!]
  \centering
   \includegraphics[width=1.0\linewidth]{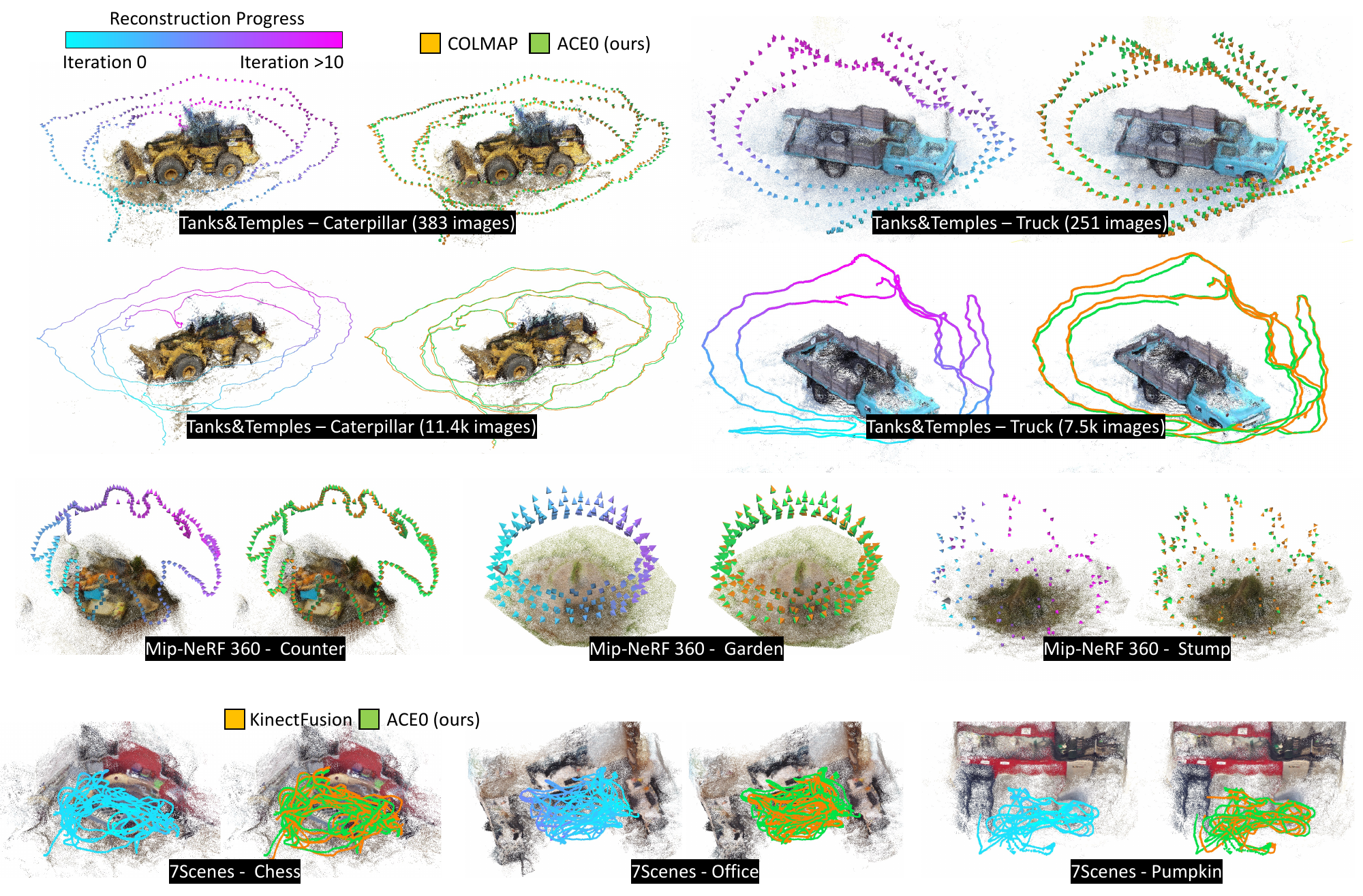}
   \caption{\textbf{More Reconstructed Poses.} We show poses estimated by ACE0. We color code the reconstruction iteration in which a particular view has been registered. We show the ACE0 point cloud as a representation of the scene. We also compare our poses to poses estimated by COLMAP (Mip-NeRF 360, Tanks and Temples) and KinectFusion (7-Scenes). }
   \label{fig:supp:exp:qualitative}
\end{figure*}

\begin{figure}[t]
  \centering
   \includegraphics[width=1.0\linewidth]{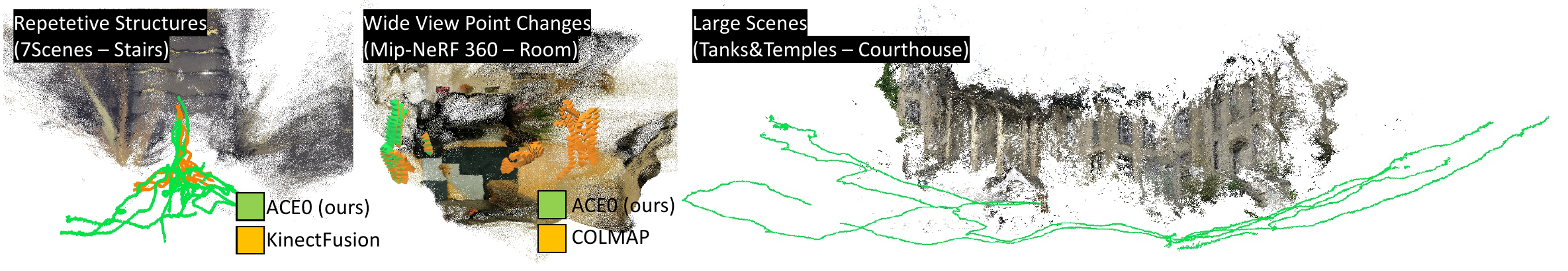}
   \caption{\textbf{Failure Cases. Stairs:} Parts of the reconstruction collapse due to visual ambiguity. \textbf{Room:} ACE0 fails to register the views on the right due to low visual overlap. \textbf{Courthouse} is too large to be represented well by a single MLP.}
   \label{fig:exp:failures}
\end{figure}

\section{Additional Results}
\label{sec:app:results}

\paragraph{More Qualitative Results.} 
We show estimated poses for more scenes in Fig.~\ref{fig:supp:exp:qualitative}.
We show failure cases in Fig.~\ref{fig:exp:failures} corresponding to limitations discussed in \Cref{sec:conclusion}.

\begin{figure*}[h!]
  \centering
   \vspace{1cm}
   \includegraphics[width=1.0\linewidth]{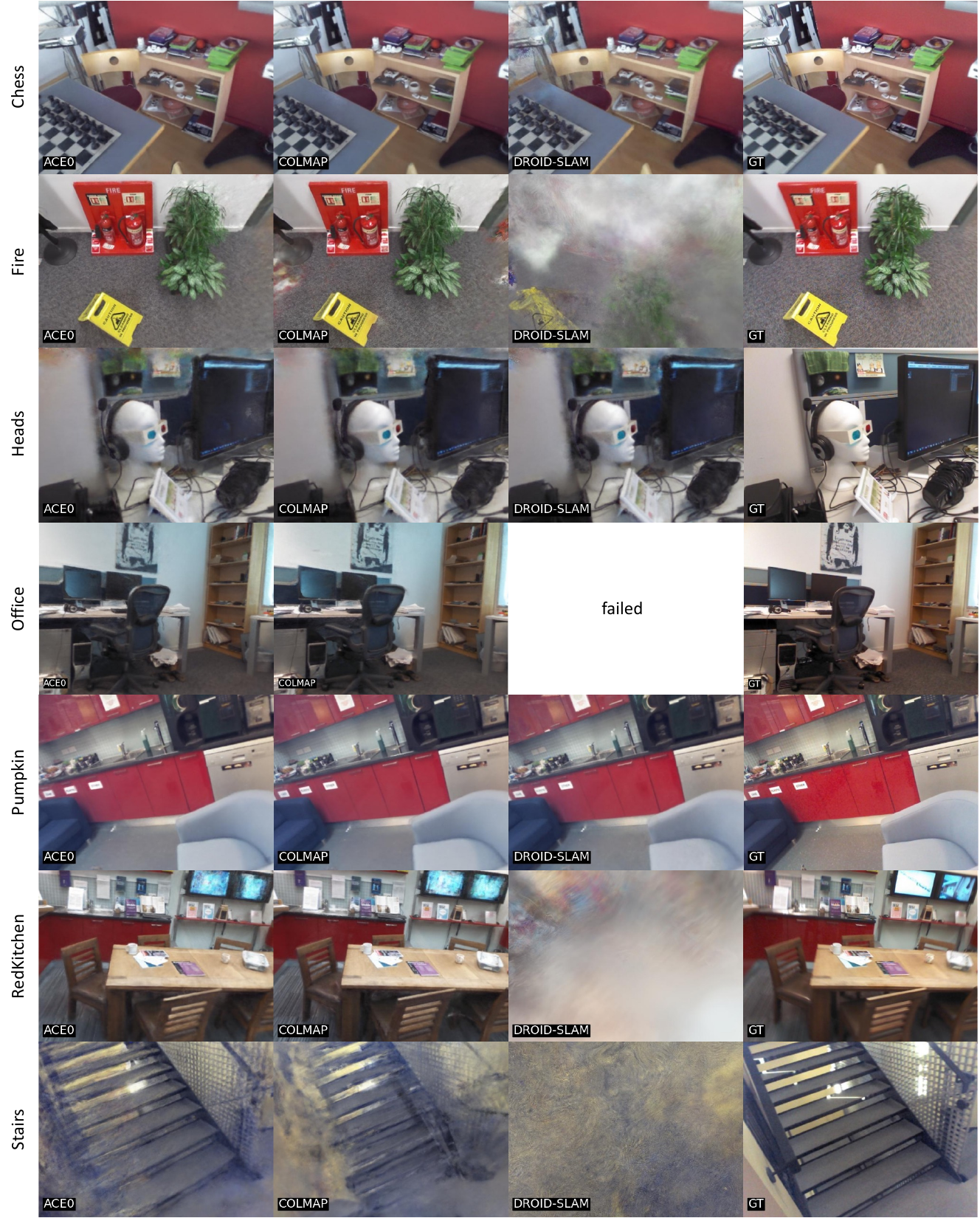}
   \caption{\textbf{View Synthesis Quality on 7-Scenes.} For each scene, we show the test image where ACE0 achieves its median PSNR value. For comparison, we show the corresponding results of our baselines as well as the true test image (GT). These results correspond to \Cref{tab:sevenscenes}. ``COLMAP'' refers to COLMAP with \emph{default} parameters.}
   \label{fig:supp:exp:renders:7s:full}
\end{figure*}

\begin{figure*}[h!]
  \centering
   \includegraphics[width=1.0\linewidth]{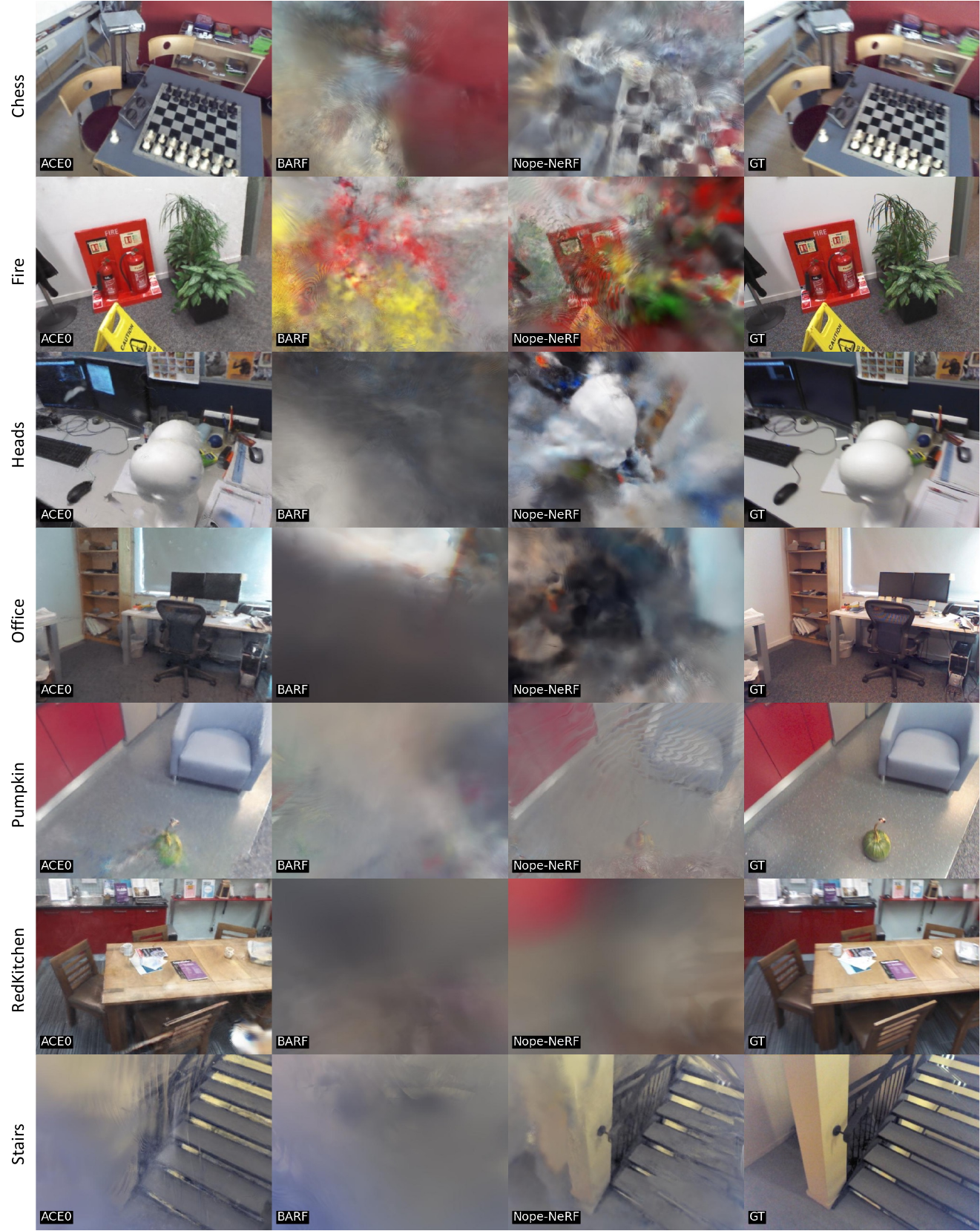}
   \caption{\textbf{View Synthesis Quality on 7-Scenes (200 Images Subset).} For each scene, we show the test image where ACE0 achieves its median PSNR value. For comparison, we show the corresponding results of our baselines as well as the true test image (GT). For all methods, we obtain these results on a subset of 200 images per scene to achieve a reasonable training time of NoPe-NeRF. These results correspond to \Cref{tab:sevenscenes}(right).}
   \label{fig:supp:exp:renders:7s:nopenerf}
\end{figure*}

\begin{figure*}[h!]
  \centering
   \includegraphics[width=0.77\linewidth]{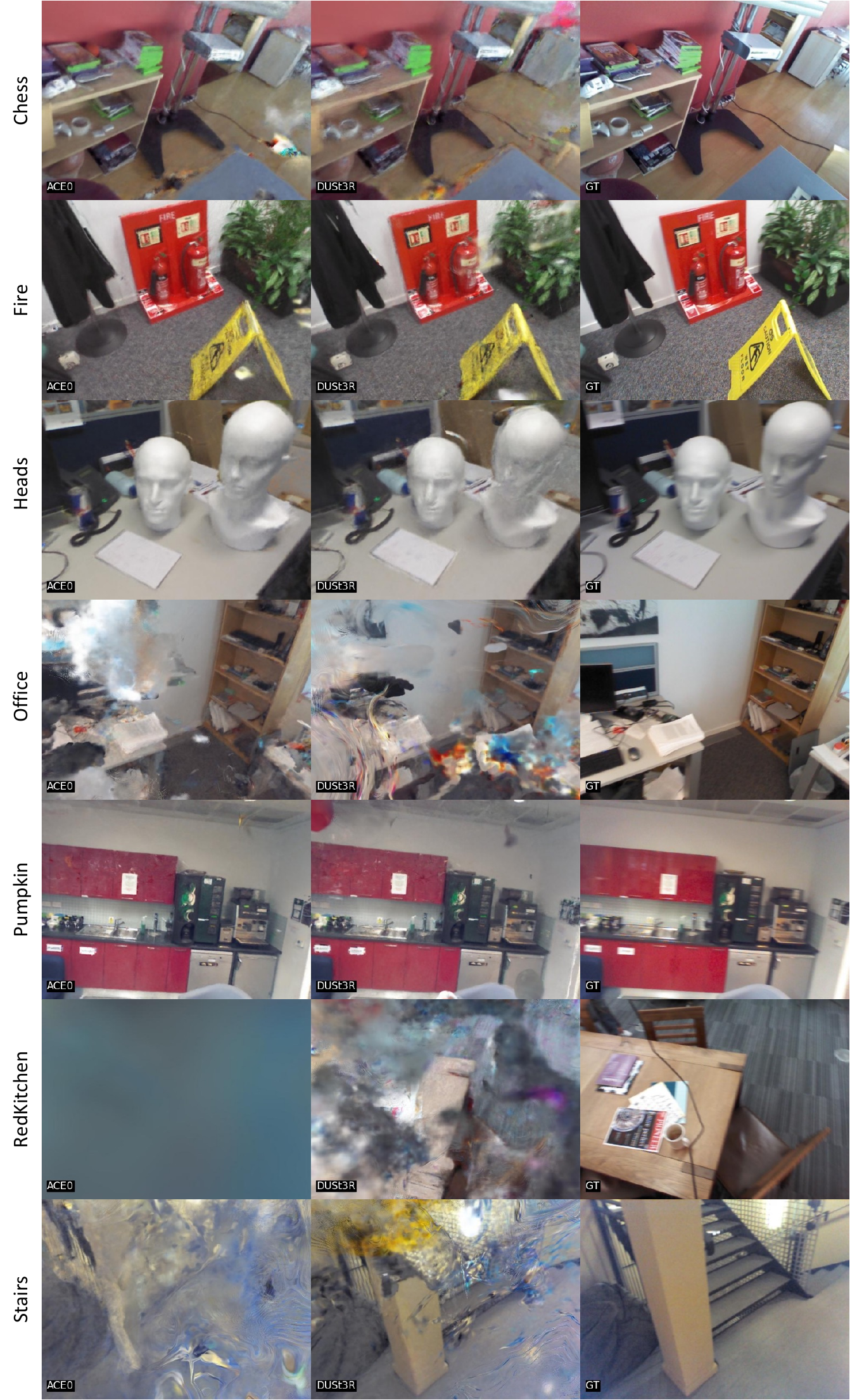}
   \caption{\textbf{View Synthesis Quality on 7-Scenes (50 Images Subset).} For each scene, we show the test image where ACE0 achieves its median PSNR value. For comparison, we show the corresponding results of our baselines as well as the true test image (GT). For all methods, we obtain these results on a subset of 50 images per scene to prevent DUSt3R from running out of GPU memory. These results correspond to \Cref{tab:sevenscenes}(right).}
   \label{fig:supp:exp:renders:7s:dust3r}
   \vspace{1cm}
\end{figure*}

\begin{figure*}[h!]
  \centering
   \includegraphics[width=1.0\linewidth]{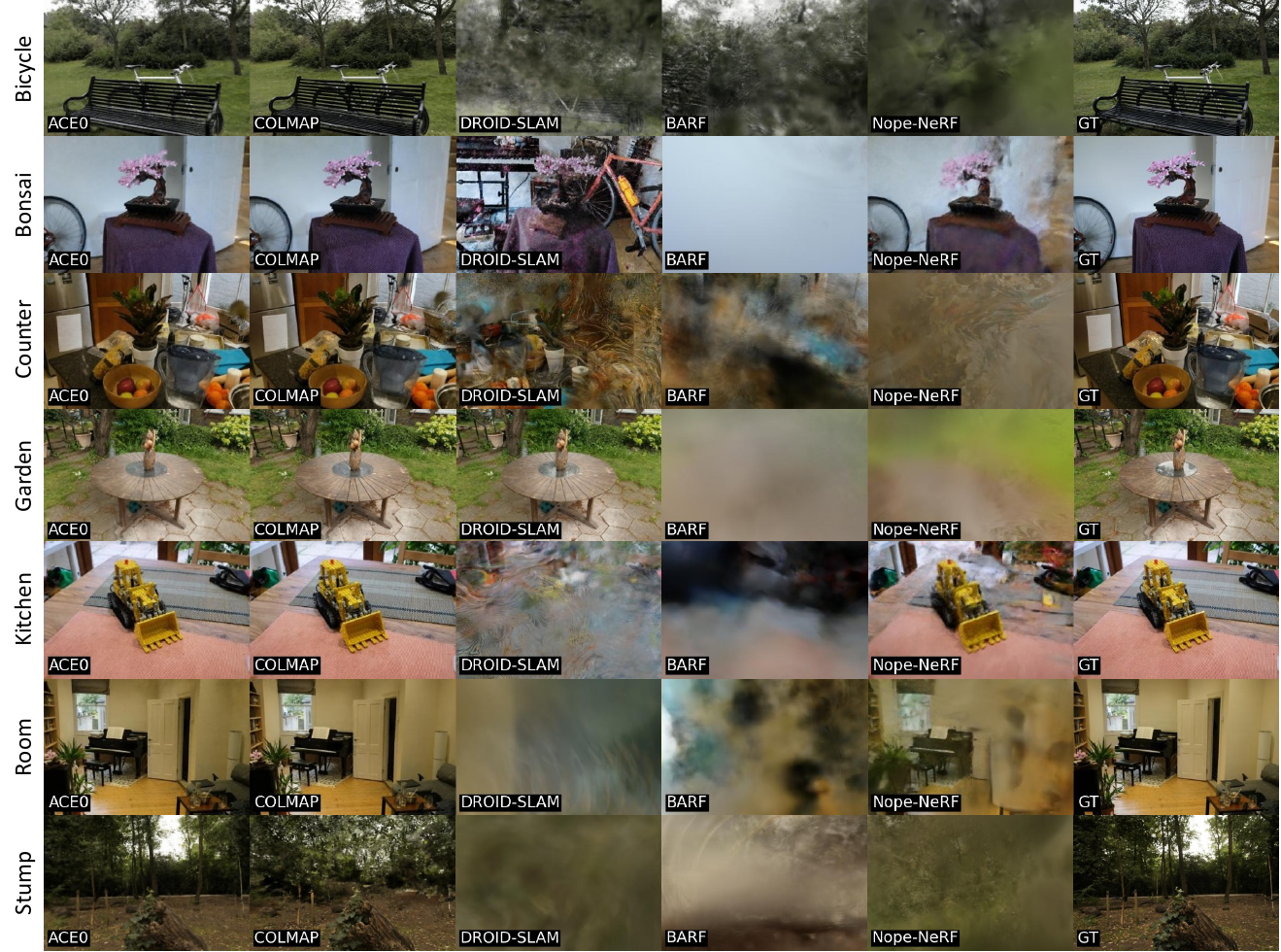}
   \caption{\textbf{View Synthesis Quality on the MIP-NeRF 360 dataset.} For each scene, we show the test image where ACE0 achieves its median PSNR value. For comparison, we show the corresponding results of our baselines as well as the true test image (GT). These results correspond to \subref{tab:mip}.}
   \label{fig:supp:exp:renders:mip360}
\end{figure*}

\begin{figure*}[h!]
  \centering
   \includegraphics[width=0.94\linewidth]{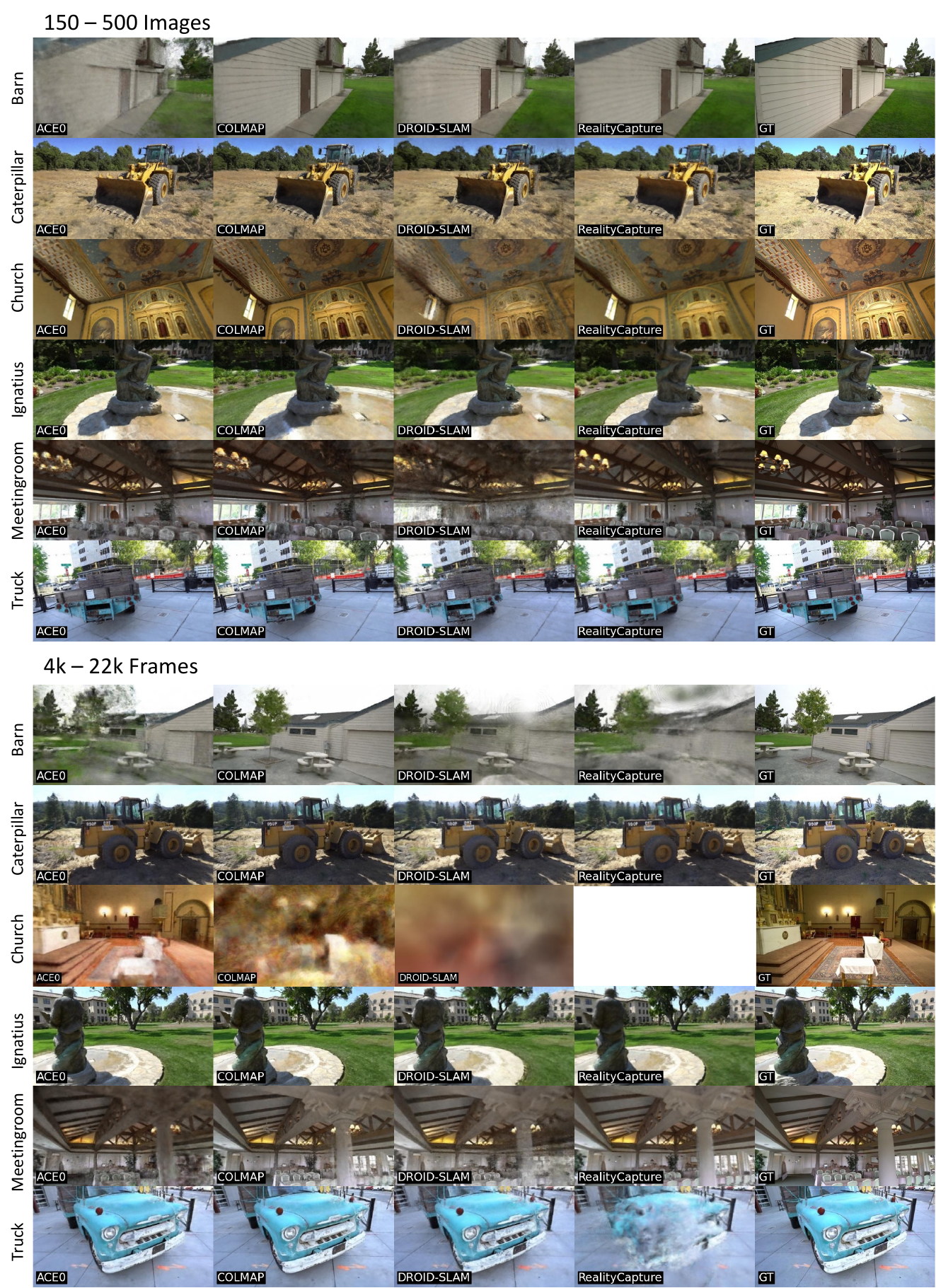}
   \caption{\textbf{View Synthesis Quality on Tanks and Temples (Training Scenes).} For each scene, we show the test image where ACE0 achieves its median PSNR value. For comparison, we show the corresponding results of our baselines as well as the true test image (GT). These results correspond to \Cref{tab:t2}. \textbf{Top:} Results based on 150-500 images per scenes. \textbf{Bottom:} Results based on 4k-22k frames sampled from the full video. }
   \label{fig:supp:exp:renders:t2:training}
\end{figure*}

\begin{figure*}[h!]
  \centering
   \includegraphics[width=0.94\linewidth]{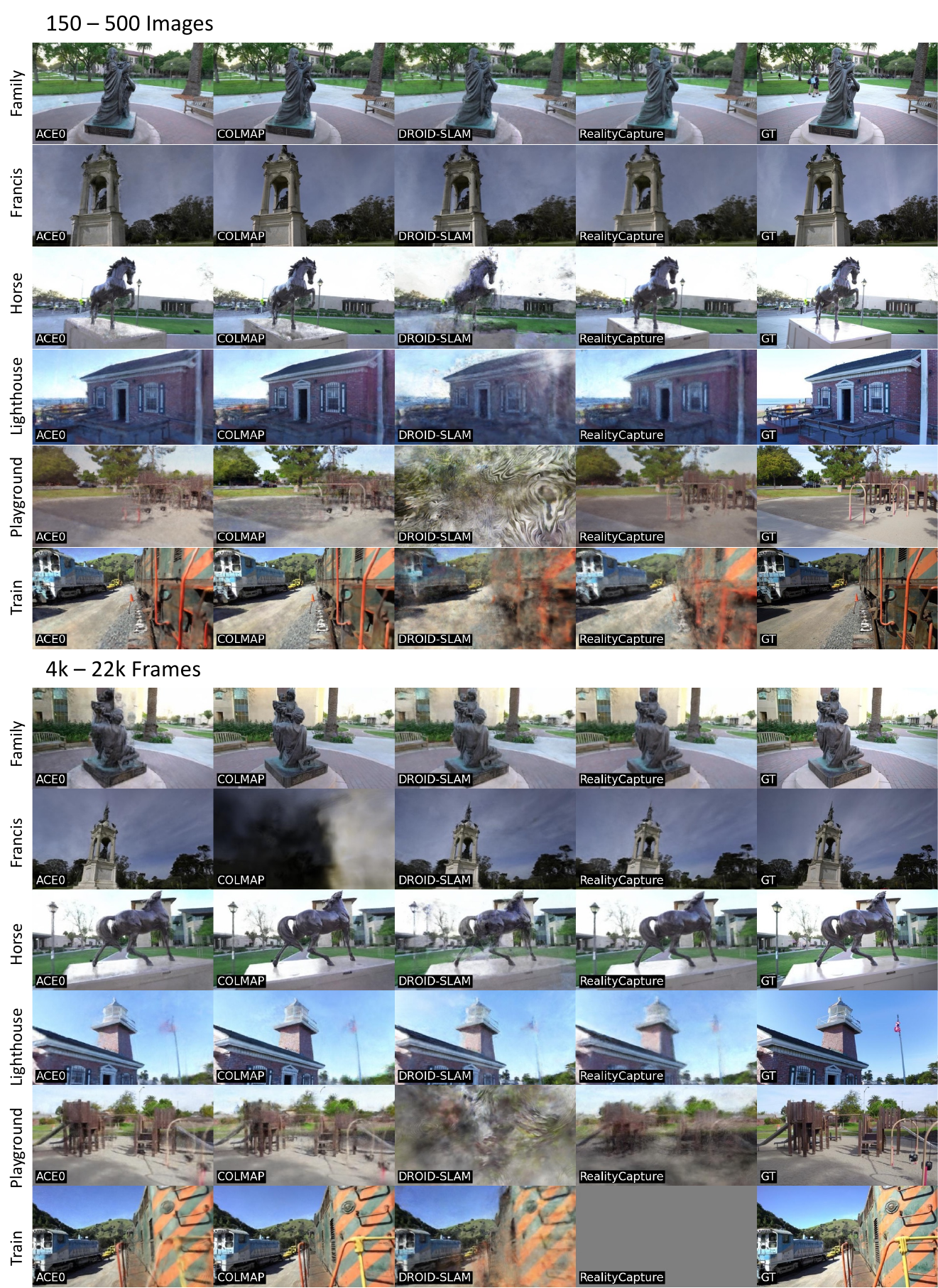}
   \caption{\textbf{View Synthesis Quality on Tanks and Temples (Intermediate Scenes).} For each scene, we show the test image where ACE0 achieves its median PSNR value. For comparison, we show the corresponding results of our baselines as well as the true test image (GT). These results correspond to \Cref{tab:t2}. \textbf{Top:} Results based on 150-500 images per scenes. \textbf{Bottom:} Results based on 4k-22k frames sampled from the full video. }
   \label{fig:supp:exp:renders:t2:intermediate}
\end{figure*}

\begin{figure*}[h!]
  \centering
   \includegraphics[width=1.0\linewidth]{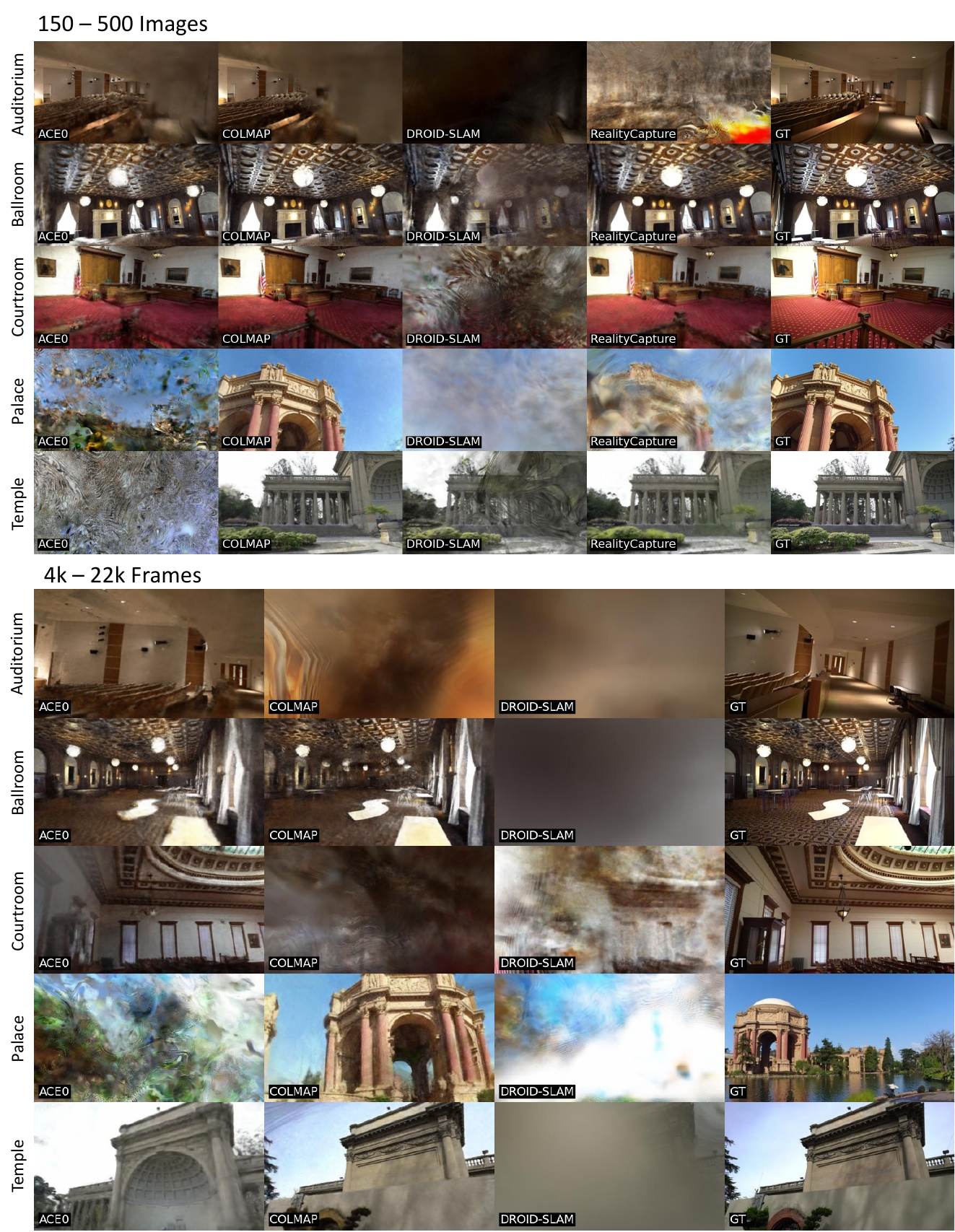}
   \caption{\textbf{View Synthesis Quality on Tanks and Temples (Advanced Scenes).} For each scene, we show the test image where ACE0 achieves its median PSNR value. For comparison, we show the corresponding results of our baselines as well as the true test image (GT). These results correspond to \Cref{tab:t2}. \textbf{Top:} Results based on 150-500 images per scenes. \textbf{Bottom:} Results based on 4k-22k frames sampled from the full video. }
   \label{fig:supp:exp:renders:t2:advanced}
\end{figure*}

Additionally, we show synthesized views based on ACE0 poses. 
Corresponding to our quantitative results (\Cref{tab:sevenscenes}, \subref{tab:mip}, \Cref{tab:t2}), we select one test image per scene that is representative of the view synthesis quality of ACE0 on that scene.
More specifically, we select the test image where ACE0 achieves its median PSNR value. 
For comparison, we synthesise the same image using the estimated poses of our competitors.
For synthesised images of 7-Scenes, see Fig.~\ref{fig:supp:exp:renders:7s:full}. 
For comparison with BARF and NoPe-NeRF, we report results on a 200 image subset per scene.
We show the associated synthesized images in Fig.~\ref{fig:supp:exp:renders:7s:nopenerf}.
For comparison with DUSt3R, we report results on a 50 image subset per scene. 
We show the associated synthesized images in Fig.~\ref{fig:supp:exp:renders:7s:dust3r}.
We show synthesized images for the MIP-NeRF 360 dataset in Fig.~\ref{fig:supp:exp:renders:mip360}.
We also show synthesized images for Tanks and Temples, in Fig.~\ref{fig:supp:exp:renders:t2:training} for \emph{Training} scenes, in Fig.~\ref{fig:supp:exp:renders:t2:intermediate} for \emph{Intermediate} scenes, and in Fig.~\ref{fig:supp:exp:renders:t2:advanced} for \emph{Advanced} scenes.

\paragraph{More Quantitative Results.} 
We augment \Cref{tab:sevenscenes} with \mbox{SSIM~\cite{wang2004image}} and \mbox{LPIPS~\cite{zhang2018perceptual}} scores for the 7-Scenes dataset, \textit{c.f.}, \Cref{tab:supp:sevenscenes_ssim,tab:supp:sevenscenes_lpips} respectively.
Similarly, \Cref{tab:supp:t2_ssim,tab:supp:t2_lpips} show \mbox{SSIM~\cite{wang2004image}} and \mbox{LPIPS~\cite{zhang2018perceptual}} scores for the Tanks and Temples dataset, augmenting \Cref{tab:t2}.
Furthermore, \Cref{tab:supp:mip_ssim,tab:supp:mip_lpips} show \mbox{SSIM~\cite{wang2004image}} and \mbox{LPIPS~\cite{zhang2018perceptual}} scores for the \mbox{Mip-NeRF~360} dataset, augmenting \subref{tab:mip}.
In all cases, SSIM and LPIPS behave very similar to PSNR in our experiments.
 
\begin{table}[]
  \adjustbox{width=\linewidth}{
            \begin{tabular}{cc||cc||cccccccccccc||ccc||cc}
             & & & & & & & & & & & & & & & & & & & & \\
            \toprule
            \multicolumn{2}{c||}{~} &%
            \multicolumn{2}{c||}{Pseudo Ground Truth} &%
            \multicolumn{12}{c||}{All Frames} &%
            \multicolumn{3}{c||}{200 Frames} &%
            \multicolumn{2}{c}{50 Frames} \\
            \hhline{~~-------------------}
             ~ &%
            \hspace{-6pt}\parbox[t]{0mm}{\multirow{2}{*}{\rotatebox[origin=c]{90}{Frames\hspace{-12pt}}}} &%
            Kinect &%
            COLMAP &%
            \multicolumn{3}{c}{COLMAP} &%
            \multicolumn{3}{c}{DROID-SLAM$^\dagger$} &%
            \multicolumn{3}{c}{ACE0} &%
            \multicolumn{3}{c||}{KF+ACE0} &%
            BARF &%
            NoPE-NeRF$^\dagger$ &%
            ACE0 &%
            DUSt3R &%
            ACE0 \\
             &%
             ~ &%
            Fusion &%
            (default) &%
            \multicolumn{3}{c}{(fast)} &%
            \multicolumn{3}{c}{\cite{teed2021droid}} &%
            \multicolumn{3}{c}{(ours)} &%
            \multicolumn{3}{c||}{(ours)} &%
            \cite{lin2021barf} &%
            \cite{bian2023nope} &%
            (ours) &%
            \cite{wang2024dust3r} &%
            (Ours) \\
            \midrule 
             Chess & 6k & 0.69 & 0.86 & \multicolumn{3}{c}{\begin{tabular}{ccc} & ~{\cellcolor{TableLightGreen}0.86}~ & \end{tabular}} &  \multicolumn{3}{c}{\begin{tabular}{ccc} & ~{\cellcolor{TableYellow}0.76}~ & \end{tabular}} &  \multicolumn{3}{c}{\begin{tabular}{ccc} & ~{\cellcolor{TableLightGreen}0.84}~ & \end{tabular}} &  \multicolumn{3}{c||}{\begin{tabular}{ccc} & ~{\cellcolor{TableLightGreen}0.84}~ & \end{tabular}} &  0.55 & 0.51 & \textbf{0.82} & 0.66 &\textbf{0.70} \\
             & & & & & & & & & & & & & & & & & & & &\\[-10pt]
 Fire & 4k & 0.55 & 0.73 & \multicolumn{3}{c}{\begin{tabular}{ccc} & ~{\cellcolor{TableLightGreen}0.73}~ & \end{tabular}} &  \multicolumn{3}{c}{\begin{tabular}{ccc} & ~{\cellcolor{TableRed}0.44}~ & \end{tabular}} &  \multicolumn{3}{c}{\begin{tabular}{ccc} & ~{\cellcolor{TableLightGreen}0.69}~ & \end{tabular}} &  \multicolumn{3}{c||}{\begin{tabular}{ccc} & ~{\cellcolor{TableLightGreen}0.70}~ & \end{tabular}} &  0.45 & 0.41 & \textbf{0.70} & 0.55 &\textbf{0.59} \\
             & & & & & & & & & & & & & & & & & & & &\\[-10pt]
 Heads & 2k & 0.69 & 0.82 & \multicolumn{3}{c}{\begin{tabular}{ccc} & ~{\cellcolor{TableLightGreen}0.82}~ & \end{tabular}} &  \multicolumn{3}{c}{\begin{tabular}{ccc} & ~{\cellcolor{TableLightGreen}0.78}~ & \end{tabular}} &  \multicolumn{3}{c}{\begin{tabular}{ccc} & ~{\cellcolor{TableLightGreen}0.81}~ & \end{tabular}} &  \multicolumn{3}{c||}{\begin{tabular}{ccc} & ~{\cellcolor{TableLightGreen}0.81}~ & \end{tabular}} &  0.54 & 0.52 & \textbf{0.80} & 0.69 &\textbf{0.79} \\
             & & & & & & & & & & & & & & & & & & & &\\[-10pt]
 Office & 10k & 0.72 & 0.85 & \multicolumn{3}{c}{\begin{tabular}{ccc} & ~{\cellcolor{TableLightGreen}0.84}~ & \end{tabular}} &  \multicolumn{3}{c}{\begin{tabular}{ccc} & {\cellcolor{TableRed}failed} & \end{tabular}} &  \multicolumn{3}{c}{\begin{tabular}{ccc} & ~{\cellcolor{TableLightGreen}0.83}~ & \end{tabular}} &  \multicolumn{3}{c||}{\begin{tabular}{ccc} & ~{\cellcolor{TableLightGreen}0.83}~ & \end{tabular}} &  0.59 & 0.55 & \textbf{0.80} & 0.57 &\textbf{0.61} \\
             & & & & & & & & & & & & & & & & & & & &\\[-10pt]
 Pumpkin & 6k & 0.71 & 0.84 & \multicolumn{3}{c}{\begin{tabular}{ccc} & ~{\cellcolor{TableLightGreen}0.84}~ & \end{tabular}} &  \multicolumn{3}{c}{\begin{tabular}{ccc} & ~{\cellcolor{TableRed}0.69}~ & \end{tabular}} &  \multicolumn{3}{c}{\begin{tabular}{ccc} & ~{\cellcolor{TableLightGreen}0.83}~ & \end{tabular}} &  \multicolumn{3}{c||}{\begin{tabular}{ccc} & ~{\cellcolor{TableLightGreen}0.82}~ & \end{tabular}} &  0.75 & 0.65 & \textbf{0.84} & 0.76 &\textbf{0.77} \\
             & & & & & & & & & & & & & & & & & & & &\\[-10pt]
 RedKitchen & 12k & 0.61 & 0.80 & \multicolumn{3}{c}{\begin{tabular}{ccc} & ~{\cellcolor{TableLightGreen}0.80}~ & \end{tabular}} &  \multicolumn{3}{c}{\begin{tabular}{ccc} & ~{\cellcolor{TableRed}0.51}~ & \end{tabular}} &  \multicolumn{3}{c}{\begin{tabular}{ccc} & ~{\cellcolor{TableLightGreen}0.77}~ & \end{tabular}} &  \multicolumn{3}{c||}{\begin{tabular}{ccc} & ~{\cellcolor{TableLightGreen}0.78}~ & \end{tabular}} &  0.52 & 0.50 & \textbf{0.72} & 0.54 &\textbf{0.55} \\
             & & & & & & & & & & & & & & & & & & & &\\[-10pt]
 Stairs & 3k & 0.63 & 0.59 & \multicolumn{3}{c}{\begin{tabular}{ccc} & ~{\cellcolor{TableDarkGreen}0.77}~ & \end{tabular}} &  \multicolumn{3}{c}{\begin{tabular}{ccc} & ~{\cellcolor{TableRed}0.47}~ & \end{tabular}} &  \multicolumn{3}{c}{\begin{tabular}{ccc} & ~{\cellcolor{TableLightGreen}0.57}~ & \end{tabular}} &  \multicolumn{3}{c||}{\begin{tabular}{ccc} & ~{\cellcolor{TableDarkGreen}0.70}~ & \end{tabular}} &  0.65 & 0.66 & \textbf{0.68} & 0.54 &\textbf{0.55} \\
             & & & & & & & & & & & & & & & & & & & &\\[-10pt]
\hhline{~--------------------}
             Average &      & 0.66 & 0.78 & \multicolumn{3}{c}{\begin{tabular}{ccc} & ~{\cellcolor{TableLightGreen}0.81}~ & \end{tabular}} &  \multicolumn{3}{c}{\begin{tabular}{ccc} & {\cellcolor{TableRed}N/A} & \end{tabular}} &  \multicolumn{3}{c}{\begin{tabular}{ccc} & ~{\cellcolor{TableLightGreen}0.76}~ & \end{tabular}} &  \multicolumn{3}{c||}{\begin{tabular}{ccc} & ~{\cellcolor{TableLightGreen}0.78}~ & \end{tabular}} &  0.58 & 0.54 & \textbf{0.76} & 0.62 &\textbf{0.65} \\
             & & & & & & & & & & & & & & & & & & & &\\[-10pt]
\hline
             Avg. Time &      & realtime & 38h & \multicolumn{3}{c}{\begin{tabular}{ccc} & ~13h~ & \end{tabular}} &  \multicolumn{3}{c}{\begin{tabular}{ccc} & ~18min~ & \end{tabular}} &  \multicolumn{3}{c}{\begin{tabular}{ccc} & ~1h~ & \end{tabular}} &  \multicolumn{3}{c||}{\begin{tabular}{ccc} & ~7min~ & \end{tabular}} &  8.5h & 47h & \textbf{27min} & \textbf{4min*} &16min \\
             & & & & & & & & & & & & & & & & & & & &\\[-10pt]
\bottomrule
      \end{tabular}
    
  }
  \caption{
      \textbf{7-Scenes - SSIM ($\uparrow$).} 
      We show the pose accuracy via view synthesis with Nerfacto~\cite{nerfstudio} as \mbox{\textbf{SSIM}~($\uparrow$)~\cite{wang2004image}}, and the reconstruction time.
      Results for \emph{All Frames} are color coded \wrt similarity to the COLMAP pseudo ground truth:
      \TCBTableDarkGreen{$>0.05$ better}, \TCBTableLightGreen{within $\pm 0.05$}, \TCBTableYellow{$>0.05$ worse}, \TCBTableRed{$>0.1$ worse}.
      For some competitors, we had to sub-sample the images due to their computational complexity (right side).
      $^{\dagger}$Method needs sequential inputs. $^{*}$Results on more powerful hardware.
  }%
  \label{tab:supp:sevenscenes_ssim}%
\end{table}

\begin{table}[]
  \adjustbox{width=\linewidth}{
            \begin{tabular}{cc||cc||cccccccccccc||ccc||cc}
             & & & & & & & & & & & & & & & & & & & & \\
            \toprule
            \multicolumn{2}{c||}{~} &%
            \multicolumn{2}{c||}{Pseudo Ground Truth} &%
            \multicolumn{12}{c||}{All Frames} &%
            \multicolumn{3}{c||}{200 Frames} &%
            \multicolumn{2}{c}{50 Frames} \\
            \hhline{~~-------------------}
             ~ &%
            \hspace{-6pt}\parbox[t]{0mm}{\multirow{2}{*}{\rotatebox[origin=c]{90}{Frames\hspace{-12pt}}}} &%
            Kinect &%
            COLMAP &%
            \multicolumn{3}{c}{COLMAP} &%
            \multicolumn{3}{c}{DROID-SLAM$^\dagger$} &%
            \multicolumn{3}{c}{ACE0} &%
            \multicolumn{3}{c||}{KF+ACE0} &%
            BARF &%
            NoPE-NeRF$^\dagger$ &%
            ACE0 &%
            DUSt3R &%
            ACE0 \\
             &%
             ~ &%
            Fusion &%
            (default) &%
            \multicolumn{3}{c}{(fast)} &%
            \multicolumn{3}{c}{\cite{teed2021droid}} &%
            \multicolumn{3}{c}{(ours)} &%
            \multicolumn{3}{c||}{(ours)} &%
            \cite{lin2021barf} &%
            \cite{bian2023nope} &%
            (ours) &%
            \cite{wang2024dust3r} &%
            (Ours) \\
            \midrule 
             Chess & 6k & 0.37 & 0.17 & \multicolumn{3}{c}{\begin{tabular}{ccc} & ~{\cellcolor{TableLightGreen}0.17}~ & \end{tabular}} &  \multicolumn{3}{c}{\begin{tabular}{ccc} & ~{\cellcolor{TableRed}0.33}~ & \end{tabular}} &  \multicolumn{3}{c}{\begin{tabular}{ccc} & ~{\cellcolor{TableLightGreen}0.18}~ & \end{tabular}} &  \multicolumn{3}{c||}{\begin{tabular}{ccc} & ~{\cellcolor{TableLightGreen}0.18}~ & \end{tabular}} &  0.80 & 0.76 & \textbf{0.20} & 0.42 &\textbf{0.36} \\
             & & & & & & & & & & & & & & & & & & & &\\[-10pt]
 Fire & 4k & 0.44 & 0.27 & \multicolumn{3}{c}{\begin{tabular}{ccc} & ~{\cellcolor{TableLightGreen}0.27}~ & \end{tabular}} &  \multicolumn{3}{c}{\begin{tabular}{ccc} & ~{\cellcolor{TableRed}0.70}~ & \end{tabular}} &  \multicolumn{3}{c}{\begin{tabular}{ccc} & ~{\cellcolor{TableLightGreen}0.29}~ & \end{tabular}} &  \multicolumn{3}{c||}{\begin{tabular}{ccc} & ~{\cellcolor{TableLightGreen}0.29}~ & \end{tabular}} &  0.79 & 0.80 & \textbf{0.27} & 0.40 &\textbf{0.37} \\
             & & & & & & & & & & & & & & & & & & & &\\[-10pt]
 Heads & 2k & 0.42 & 0.26 & \multicolumn{3}{c}{\begin{tabular}{ccc} & ~{\cellcolor{TableLightGreen}0.26}~ & \end{tabular}} &  \multicolumn{3}{c}{\begin{tabular}{ccc} & ~{\cellcolor{TableYellow}0.32}~ & \end{tabular}} &  \multicolumn{3}{c}{\begin{tabular}{ccc} & ~{\cellcolor{TableLightGreen}0.26}~ & \end{tabular}} &  \multicolumn{3}{c||}{\begin{tabular}{ccc} & ~{\cellcolor{TableLightGreen}0.27}~ & \end{tabular}} &  0.75 & 0.70 & \textbf{0.28} & 0.40 &\textbf{0.28} \\
             & & & & & & & & & & & & & & & & & & & &\\[-10pt]
 Office & 10k & 0.42 & 0.25 & \multicolumn{3}{c}{\begin{tabular}{ccc} & ~{\cellcolor{TableLightGreen}0.24}~ & \end{tabular}} &  \multicolumn{3}{c}{\begin{tabular}{ccc} & {\cellcolor{TableRed}failed} & \end{tabular}} &  \multicolumn{3}{c}{\begin{tabular}{ccc} & ~{\cellcolor{TableLightGreen}0.27}~ & \end{tabular}} &  \multicolumn{3}{c||}{\begin{tabular}{ccc} & ~{\cellcolor{TableLightGreen}0.26}~ & \end{tabular}} &  0.77 & 0.76 & \textbf{0.31} & 0.64 &\textbf{0.54} \\
             & & & & & & & & & & & & & & & & & & & &\\[-10pt]
 Pumpkin & 6k & 0.40 & 0.23 & \multicolumn{3}{c}{\begin{tabular}{ccc} & ~{\cellcolor{TableLightGreen}0.24}~ & \end{tabular}} &  \multicolumn{3}{c}{\begin{tabular}{ccc} & ~{\cellcolor{TableRed}0.50}~ & \end{tabular}} &  \multicolumn{3}{c}{\begin{tabular}{ccc} & ~{\cellcolor{TableLightGreen}0.24}~ & \end{tabular}} &  \multicolumn{3}{c||}{\begin{tabular}{ccc} & ~{\cellcolor{TableLightGreen}0.25}~ & \end{tabular}} &  0.45 & 0.74 & \textbf{0.22} & 0.30 &\textbf{0.27} \\
             & & & & & & & & & & & & & & & & & & & &\\[-10pt]
 RedKitchen & 12k & 0.51 & 0.25 & \multicolumn{3}{c}{\begin{tabular}{ccc} & ~{\cellcolor{TableLightGreen}0.25}~ & \end{tabular}} &  \multicolumn{3}{c}{\begin{tabular}{ccc} & ~{\cellcolor{TableRed}0.82}~ & \end{tabular}} &  \multicolumn{3}{c}{\begin{tabular}{ccc} & ~{\cellcolor{TableLightGreen}0.27}~ & \end{tabular}} &  \multicolumn{3}{c||}{\begin{tabular}{ccc} & ~{\cellcolor{TableLightGreen}0.27}~ & \end{tabular}} &  0.84 & 0.84 & \textbf{0.35} & \textbf{0.62} &0.70 \\
             & & & & & & & & & & & & & & & & & & & &\\[-10pt]
 Stairs & 3k & 0.39 & 0.50 & \multicolumn{3}{c}{\begin{tabular}{ccc} & ~{\cellcolor{TableDarkGreen}0.30}~ & \end{tabular}} &  \multicolumn{3}{c}{\begin{tabular}{ccc} & ~{\cellcolor{TableRed}0.83}~ & \end{tabular}} &  \multicolumn{3}{c}{\begin{tabular}{ccc} & ~{\cellcolor{TableLightGreen}0.51}~ & \end{tabular}} &  \multicolumn{3}{c||}{\begin{tabular}{ccc} & ~{\cellcolor{TableDarkGreen}0.34}~ & \end{tabular}} &  0.68 & 0.70 & \textbf{0.43} & 0.74 &\textbf{0.61} \\
             & & & & & & & & & & & & & & & & & & & &\\[-10pt]
\hhline{~--------------------}
             Average &      & 0.42 & 0.28 & \multicolumn{3}{c}{\begin{tabular}{ccc} & ~{\cellcolor{TableLightGreen}0.25}~ & \end{tabular}} &  \multicolumn{3}{c}{\begin{tabular}{ccc} & {\cellcolor{TableRed}N/A} & \end{tabular}} &  \multicolumn{3}{c}{\begin{tabular}{ccc} & ~{\cellcolor{TableLightGreen}0.29}~ & \end{tabular}} &  \multicolumn{3}{c||}{\begin{tabular}{ccc} & ~{\cellcolor{TableLightGreen}0.26}~ & \end{tabular}} &  0.73 & 0.76 & \textbf{0.29} & 0.50 &\textbf{0.45} \\
             & & & & & & & & & & & & & & & & & & & &\\[-10pt]
\hline
             Avg. Time &      & realtime & 38h & \multicolumn{3}{c}{\begin{tabular}{ccc} & ~13h~ & \end{tabular}} &  \multicolumn{3}{c}{\begin{tabular}{ccc} & ~18min~ & \end{tabular}} &  \multicolumn{3}{c}{\begin{tabular}{ccc} & ~1h~ & \end{tabular}} &  \multicolumn{3}{c||}{\begin{tabular}{ccc} & ~7min~ & \end{tabular}} &  8.5h & 47h & \textbf{27min} & \textbf{4min*} &16min \\
             & & & & & & & & & & & & & & & & & & & &\\[-10pt]
\bottomrule
      \end{tabular}
    
  }
  \caption{
      \textbf{7-Scenes - LPIPS ($\downarrow$).} 
      We show the pose accuracy via view synthesis with Nerfacto~\cite{nerfstudio} as \mbox{\textbf{LPIPS}~($\downarrow$)~\cite{zhang2018perceptual}}, and the reconstruction time.
      Results for \emph{All Frames} are color coded \wrt similarity to the COLMAP pseudo ground truth:
      \TCBTableDarkGreen{$>0.05$ better (lower)}, \TCBTableLightGreen{within $\pm 0.05$}, \TCBTableYellow{$>0.05$ worse (higher)}, \TCBTableRed{$>0.1$ worse (higher)}.
      For some competitors, we had to sub-sample the images due to their computational complexity (right side).
      $^{\dagger}$Method needs sequential inputs. $^{*}$Results on more powerful hardware.
  }%
  \label{tab:supp:sevenscenes_lpips}%
\end{table}

\paragraph{Analysis.}
We show variations of ACE0 in Table \ref{tab:supp:analysis} alongside PSNR numbers and reconstruction times for 7-Scenes.
ACE0 picks 5 random seed images to start the reconstruction. 
We show average statistics of ACE0 over five runs, starting off with different sets of seed images each time.
The standard deviation is low with 0.2 dB in PSNR, and about 4 minutes in terms of reconstruction time.

We pair ACE0 with different depth estimators, namely ZoeDepth \cite{bhat2023zoedepth} and PlaneRCNN \cite{liu2019planercnn} but observe only small differences.
The depth estimates are only used for the seed iteration, and their impact fades throughout the reconstruction process.
Indeed, using ground truth depth, measured by a Kinect sensor, yields no noteworthy advantage either. 

We run a version of ACE0 that ingests ZoeDepth estimates for all frames and uses them in all reconstruction iterations.
Thus, rather than optimizing the reprojection error of \mbox{Eq.~\ref{eq:repro_objective}}, we optimize the Euclidean distance to pseudo ground truth scene coordinates computed from depth estimates.
Accuracy is slightly lower, and reconstruction times longer. 
The depth estimates are not multi-view consistent, and the model has problems to converge.

Omitting pose refinement during neural mapping leads to a drop in PSNR of more than 2 dB.
Refinement via an MLP that predicts updates has a small but noticeable advantage over direct optimization of poses via back-propagation. 

Finally, PNSR numbers with early stopping are similar to using the static ACE training schedule, but reconstruction times are shorter.

\paragraph{More Scenes.}
As mentioned in the main paper, we removed two scenes from the Tanks and Temples dataset because the COLMAP baseline was not able to reconstruct them after days of processing or ran out of memory.
This concerns the Courthouse and the Museum scenes in the variation with 4k frames and more.
In Table \ref{tab:supp:additional} we show ACE0 results for these two scenes, COLMAP results when using only a few hundred images, as well as other baselines.

\paragraph{Registration Rates.}
Reconstruction algorithms do not always succeed in registering all input frames to the reconstruction.
While high registration rates are in general desirable, it is also disadvantageous to register images incorrectly. 
Clearly, an algorithm that always returns the identity pose has a 100\% registration rate but is not very useful.
Therefore, in our main experimental results, we compare algorithms based on PSNR numbers rather than registration rates. 
The way we calculate PSNR numbers punishes an algorithm for any incorrect estimate whether its considered ``registered'' or not, see Sec.~\ref{sec:app:benchmark} for details.
Nevertheless, we provide registration rates of ACE0 and COLMAP in Tables \ref{tab:supp:reg_7s_mip} and \ref{tab:supp:reg_t2}.
Both algorithms achieve high registration rates for many scenes.
Scenes with low registration rates also show low PSNR numbers in our main experimental results.

\begin{table}[]
\centering
\begin{tabular}{@{}lrr@{}}
\toprule
                      & \multicolumn{1}{c}{PSNR (dB)} & \multicolumn{1}{c}{Time (min)} \\ \midrule
Stability             & \multicolumn{1}{l}{}         & \multicolumn{1}{l}{}          \\ \cmidrule(r){1-1}
Main Run              & 21.2                         & 54                          \\
5 Randomized Runs     & 21.3$\pm$0.2                 & 52$\pm$4                    \\ \midrule
Depth Estimator       &                              &                               \\ \cmidrule(r){1-1}
ZoeDepth (default)    & 21.2                & 54                 \\
PlaneRCNN             & 21.0                         & \textbf{49}                          \\
Ground Truth (Kinect) & \textbf{21.3}                         & 58                          \\ \midrule
Depth Usage           &                              &                               \\ \cmidrule(r){1-1}
Seed only (default)   & \textbf{21.2}                & \textbf{54}                 \\
All Iterations        & 20.9                         & 79                          \\ \midrule
Pose Refinement            &                              &                               \\ \cmidrule(r){1-1}
MLP (default)         & \textbf{21.2}                & \textbf{54}                 \\
Direct Pose Updates   & 20.7                         & 61                          \\
No Refinement         & 18.9                         & 63                          \\ \midrule
Early Stopping        &                              &                               \\ \cmidrule(r){1-1}
with (default)        & 21.2                & \textbf{54}                 \\
without               & \textbf{21.3}                         & 64                          \\ \bottomrule
\end{tabular}
\caption{\textbf{ACE0 Variations.} We demonstrate the impact of various design and parameter choices as PSNR in dB and reconstruction time in minutes, on 7-Scenes.}
\label{tab:supp:analysis}
\end{table}

\begin{table}[]
\centering
\begin{tabular}{@{}lccccccc@{}}
\toprule
           & \multicolumn{3}{c}{150-500 Images}                                                                                 &  & \multicolumn{3}{c}{4k-22k Frames}                                                                                            \\ \cmidrule(lr){2-4} \cmidrule(l){6-8} 
           & \begin{tabular}[c]{@{}c@{}}COLMAP\\ (default)\end{tabular}  & \begin{tabular}[c]{@{}c@{}}Reality\\Capture\end{tabular}  & \begin{tabular}[c]{@{}c@{}}ACE0\\ (ours)\end{tabular} 
           &  & \begin{tabular}[c]{@{}c@{}}Reality\\Capture\end{tabular} & \begin{tabular}[c]{@{}c@{}}ACE0\\ (ours)\end{tabular} & \begin{tabular}[c]{@{}c@{}}Sparse COLMAP\\ +ACE0 (ours)\end{tabular} \\ \midrule
Courthouse & 18.2                                                       & 15.5 & 14.4                                                  &  & 13.2                                                   & 12.5                                                  & 17.2                                                                 \\
Museum     & 17.0                                                       & 16.2 & 13.0                                                  &  & - & 14.5                                                  & 17.2                                                                 \\ \bottomrule
\end{tabular}
\caption{\textbf{Additional ACE0 Results.} We provide results for two additional scenes of Tanks and Temples where the COLMAP baseline did not finish the reconstruction (with 4k+ frames) after running for more than 5 days or ran out of memory.}
\label{tab:supp:additional}
\end{table}

\begin{table}[]
\centering
\begin{tabular}{@{}lcclcc@{}}
\toprule
7-Scenes   & \begin{tabular}[c]{@{}c@{}}ACE0\\ (ours)\end{tabular} & \begin{tabular}[c]{@{}c@{}}COLMAP\\ (``fast'')\end{tabular} &   \begin{tabular}[c]{@{}l@{}}Mip-NeRF \\ 360\end{tabular} & \begin{tabular}[c]{@{}c@{}}ACE0\\ (ours)\end{tabular} & \begin{tabular}[c]{@{}c@{}}COLMAP\\ (default)\end{tabular} \\ \midrule
Chess     & 100\%                                                 & 100\%                                                     &   Bicycle                                                 & 97.9\%                                                & 100\%                                                      \\
Fire      & 100\%                                                 & 100\%                                                     &   Bonsai                                                  & 100\%                                                 & 100\%                                                      \\
Heads     & 100\%                                                 & 100\%                                                     &   Counter                                                 & 100\%                                                 & 100\%                                                      \\
Office    & 100\%                                                 & 100\%                                                     &   Garden                                                  & 100\%                                                 & 100\%                                                      \\
Pumpkin   & 100\%                                                 & 100\%                                                     &   Kitchen                                                 & 100\%                                                 & 100\%                                                      \\
R.Kitchen & 98.4\%                                                & 100\%                                                     &   Room                                                    & 50.8\%                                                & 100\%                                                      \\
Stairs    & 100\%                                                 & 100\%                                                     &   Stump                                                   & 96.0\%                                                & 100\%                                                      \\ \bottomrule
\end{tabular}
\caption{\textbf{Registration Rates on 7-Scenes and Mip-NeRF 360.} We show the percentage of registered images for ACE0 and COLMAP. Note that the way we calculate PSNR numbers already accounts for registration rates below 100\%.}
\label{tab:supp:reg_7s_mip}
\end{table}

\begin{table*}[]
\centering
\begin{tabular}{@{}llcccccc@{}}
\toprule
                              &             & \multicolumn{2}{c}{150-500 Images}                                                                                 &  & \multicolumn{3}{c}{4k-22k Frames}                                                                                                                                                        \\ \cmidrule(lr){3-4} \cmidrule(l){6-8} 
                              &             & \begin{tabular}[c]{@{}c@{}}ACE0\\ (ours)\end{tabular} & \begin{tabular}[c]{@{}c@{}}COLMAP\\ (default)\end{tabular} &  & \begin{tabular}[c]{@{}c@{}}ACE0\\ (ours)\end{tabular} & \begin{tabular}[c]{@{}c@{}}COLMAP\\ (``fast'')\end{tabular} & \begin{tabular}[c]{@{}c@{}}Sparse COLMAP\\ +ACE0 (ours)\end{tabular} \\ \midrule
\multirow{6}{*}{\rotatebox[origin=c]{90}{Training}}     & Barn        & 94.6\%                                                & 100\%                                                      &  & 96.3\%                                                & 100\%                                                     & 100\%                                                                \\
                              & Caterpillar & 100\%                                                 & 100\%                                                      &  & 100\%                                                 & 100\%                                                     & 100\%                                                                \\
                              & Church      & 95.5\%                                                & 100\%                                                      &  & 90.3\%                                                & 99.2\%                                                    & 91.2\%                                                               \\
                              & Ignatius    & 100\%                                                 & 100\%                                                      &  & 100\%                                                 & 100\%                                                     & 100\%                                                                \\
                              & Meetingroom & 100\%                                                & 100\%                                                      &  & 94.8\%                                                 & 100\%                                                     & 99.2\%                                                                \\
                              & Truck       & 100\%                                                 & 100\%                                                      &  & 100\%                                                 & 100\%                                                     & 100\%                                                                \\ \midrule
\multirow{6}{*}{\rotatebox[origin=c]{90}{Intermediate}} & Family      & 100\%                                                 & 100\%                                                      &  & 100\%                                                 & 100\%                                                     & 100\%                                                                \\
                              & Francis     & 99.7\%                                                 & 100\%                                                      &  & 100\%                                                 & 100\%                                                     & 100\%                                                                \\
                              & Horse       & 100\%                                                 & 100\%                                                      &  & 100\%                                                 & 100\%                                                     & 100\%                                                                \\
                              & Lighthouse  & 94.5\%                                                & 100\%                                                      &  & 93.3\%                                                & 100\%                                                     & 97.8\%                                                               \\
                              & Playground  & 100\%                                                 & 100\%                                                      &  & 99.6\%                                                & 100\%                                                     & 100\%                                                                \\
                              & Train       & 96.3\%                                                & 100\%                                                      &  & 96.4\%                                                & 100\%                                                     & 97.9\%                                                               \\ \midrule
\multirow{5}{*}{\rotatebox[origin=c]{90}{Advanced}}     & Auditorium  & 98.0\%                                                & 98.7\%                                                     &  & 98.7\%                                                & 99\%                                                      & 98.3\%                                                               \\
                              & Ballroom    & 100\%                                                & 100\%                                                      &  & 100\%                                                 & 100\%                                                     & 97.8\%                                                               \\
                              & Courtroom   & 96.3\%                                                & 100\%                                                      &  & 93.1\%                                                & 100\%                                                     & 97.8\%                                                               \\
                              & Palace      & 25.7\%                                                & 99\%                                                       &  & 56.0\%                                                & 74.4\%                                                    & 77.2\%                                                               \\
                              & Temple      & 7.3\%                                                 & 100\%                                                      &  & 74.4\%                                                & 100\%                                                     & 92.4\%                                                               \\ \bottomrule
\end{tabular}
\caption{\textbf{Registration Rates on Tanks and Temples.} We show the percentage of registered images for ACE0, COLMAP, and the combination of ACE0 and COLMAP. Note that the way we calculate PSNR numbers already accounts for registration rates below 100\%.}
\label{tab:supp:reg_t2}
\end{table*}

\section{Baselines and Competitors}
\label{sec:app:baselines}

\paragraph{COLMAP.}
We run COLMAP in three variations: 
\emph{Default}, \emph{fast}, and \colmapreloc. \emph{Fast} uses parameters recommended for large image collections beyond \mbox{1\hspace{1pt}000} images \cite{fastcolmap}.
\colmapreloc uses \emph{default} parameters to obtain the initial reconstruction, and \emph{fast} parameters to register the remaining frames.
We provide the parameters of the \texttt{mapper} of the \emph{default} variation in \Cref{tab:colmap_mapper_default}.
For the \emph{fast} version, we provide the parameters that have been changed compared to \emph{default} in \Cref{tab:supp:colmap_mapper_fast}.
We also include the parameters of the feature extractor and the matcher shared by both variations in \Cref{tab:colmap_feature_extractor,tab:supp:colmap_feature_matcher}. 

As additional comparison to further optimize for efficiency, for the Tanks and Temples dataset we also computed in the second to last column of \Cref{tab:t2,tab:supp:t2_ssim,tab:supp:t2_lpips} a COLMAP variant \colmapreloc.
As presented at the end of \Cref{sec:experiments:t2}, this variant builds on the sparse reconstruction results, and performs the following steps:
\begin{enumerate}
    \item (Custom database surgery to prefix the old image names from the ``Sparse COLMAP'' model to avoid name clashes.)
    \item Feature extraction (\texttt{colmap feature\_extractor ...}) of the new images using the same parameters as COLMAP \emph{default} and \emph{fast}, \cf, \Cref{tab:colmap_feature_extractor}.
    \item Vocabulary tree matching (\texttt{colmap vocab\_tree\_matcher}) of the new images to the old images and each other, using the same parameters as COLMAP \emph{default} and \emph{fast}, \cf, \Cref{tab:supp:colmap_feature_matcher}.
    \item Image registration (\texttt{colmap image\_registrator ...}) of the new images to the old images and each other, given the matches computed in the previous step.
    The parameters for this step (\cf, \Cref{tab:supp:colmap_image_registrator}) were set to match the \texttt{mapper} parameters used in COLMAP \emph{fast}, (\cf, \Cref{tab:supp:colmap_mapper_fast}).
    \item A final BA step (\texttt{colmap bundle\_adjuster ...}) to refine the poses of the whole scene. Parameters were set similarly to COLMAP \emph{fast}, \cf, \Cref{tab:supp:colmap_bundle_adjuster}.
\end{enumerate}
Note, such an algorithm heavily relies on the fact, that the sparse set of images has been selected so that it represents the entire scene, and would need further iterative optimization for datasets where the spatial distribution of images is not known beforehand (\eg, not sequential data or data without priors from other sensors, such as GPS or WiFi).

\paragraph{NoPe-NeRF.}
We estimate poses with NoPe-NeRF by using the official code at \url{https://github.com/ActiveVisionLab/nope-nerf}.
We modify the code to ingest both training and test views, since we aim at estimating the poses of all images jointly.

For Mip-NeRF 360 scenes, we use all images per each scene reconstruction. 
As NoPe-NeRF assumes ordered image sequence, we sorted the images by name for training.

For 7-Scenes, using all frames is prohibitively expensive. 
For example, the Chess scene has 6000 frames and the implementation just freezes when using them all. 
The majority of experiments in the NoPe-NeRF paper were done with a few hundred frames, so we subsampled the 7-Scenes scans to extract a total of 200 frames per each scene.

\paragraph{DUSt3R.}
We follow the official code at \url{https://github.com/naver/dust3r} to reconstruct scenes within the 7-Scenes dataset from uncalibrated and unposed cameras.
When comparing to DUSt3R, we run ACE0 \emph{without} the initialisation from a standard Kinect v1 focal length.
That is, like DUSt3R, ACE0 estimates the focal length from scratch in the last column of \Cref{tab:sevenscenes}.
The DUSt3R framework cannot process all available images of a scene due to GPU memory constraints. 
By following the guidelines from the official repository, we configured the system to utilize a sliding window approach (w=3), enabling the accommodation of 50 frames within the 16GB memory capacity of a \mbox{NVIDIA\textsuperscript{\textregistered}} V100~\cite{nvidiav100} GPU. 
As a second version, we attempted to reconstruct 50 frames using a \mbox{NVIDIA\textsuperscript{\textregistered}} A100-40GB GPU in conjunction with the leaner 224x224 pre-trained model and an exhaustive matching configuration.
We were not able to pair the full 512x512 model with exhaustive matching, even on a A100-40GB.
Of the two configurations that fit into memory, the second one yields the most accurate poses under our experimental settings. 
Thus, we report the superior results in \Cref{tab:sevenscenes}. For further insight, we have compiled the experimental statistics in Table \ref{tab:supp:dust3r_7s}.

\begin{table}[]
\centering
\begin{tabular}{lcc}
\toprule
7-Scenes   & DUSt3R$_{512}$  & DUSt3R$_{224}$ \\ \midrule
Chess      & 16.2            & \textbf{18.9}  \\
Fire       & 13.4            & \textbf{18.8}  \\
Heads      & 13.4            & \textbf{18.4}  \\
Office     & 10.2            & \textbf{12.5}  \\
Pumpkin    & 21.0            & \textbf{21.7}  \\
R.Kitchen  & 11.5            & \textbf{13.8}  \\
Stairs     & \textbf{16.3}   & 15.3           \\ \midrule
Config.    & S.W.            & E.M.           \\ 
Hardware   & V100            & A100           \\
Memory     & 16GB            & 40GB           \\
Recon. Time& 3min            & 4min        \\
\bottomrule
\end{tabular}
\caption{\textbf{DUSt3R on 7-Scenes} S.W. denotes the sliding window configuration (window=3). E.M. denotes the exhaustive matching configuration. The performance is evaluated in PSNR (dB) using the same experiments described in \Cref{sec:experiments:sevenscenes}. In the bottom part of the table, we show the configuration, GPU hardware, GPU memory, and reconstruction time for 50 frames on each scene.}
\label{tab:supp:dust3r_7s}
\end{table}

\paragraph{DROID-SLAM.}
We estimated poses for the 7-Scenes (using only the RGB images), Tanks and Temples, and Mip-NeRF 360 datasets using the official code from: \url{https://github.com/princeton-vl/DROID-SLAM}.
In our experiments we used the same parameters that the DROID-SLAM authors chose to evaluate the method on the ETH-3D dataset and resized the input images to a resolution having an area of $384 \times 512$ pixels, while preserving the aspect ratio.
As the SLAM method requires sequential input, we sorted the images by name, although that does not remove all jumps from the datasets.
For example, each scene in the 7-Scenes dataset is composed of a sequence of disjoint scans observing the same area, and the algorithm might have difficulties in tracking accurately around the discontinuities.

\paragraph{Reality Capture\textsuperscript{\textregistered}.} We also reconstructed scenes from the Tanks and Temples dataset using Reality Capture\textsuperscript{\textregistered}, as discussed in \Cref{sec:experiments:t2}. 
For completeness, we also computed \mbox{SSIM~\cite{wang2004image}} and \mbox{LPIPS~\cite{zhang2018perceptual}} scores in \Cref{tab:supp:t2_ssim,tab:supp:t2_lpips} respectively.
On the sparse images (150-500), Reality Capture performs one order of magnitude faster than COLMAP but with slightly worse pose quality.
When we ran it on Tanks and Temples with thousands of frames, Reality Capture produced a high number of disconnected components.
We use the largest component as basis for our evaluation and regard all other frames as missing, as described in \Cref{sec:app:benchmark}.

\begin{table}
    \centering
    \begin{tabular}{@{}ll@{}}
      \toprule
      Name & value \\
      \midrule
      Max features per mpx & \num{10000} \\
      Max features per image & \num{40000} \\
      Image overlap & Medium \\
      Image downscale factor & 1 \\
      Max feature reprojection error & 2.0 \\
      \textbf{Force component rematch} & \textbf{Yes} \\ 
      Background feature detection & No \\
      Background thread priority & Low \\
      Preselector features & \num{10000} \\
      Detector sensitivity & Medium \\
      Distortion model & Brown3 \\
      \bottomrule
    \end{tabular}
    \caption{\textbf{Reality Capture alignment settings.} We used the default settings for our experiments, except for settings in \textbf{bold}.}
    \label{tab:supp:rc_alignment_settings}
\end{table}

We used Reality Capture version \texttt{1.3.2.117357} with the default alignment parameters\footnote{\url{https://rchelp.capturingreality.com/en-US/tutorials/setkeyvaluetable.htm}}, \cf, \Cref{tab:supp:rc_alignment_settings}.

The runtimes in \Cref{tab:t2,tab:supp:t2_ssim,tab:supp:t2_lpips} were measured for the iv) \emph{alignment} step only, \textbf{excluding} the steps i) \emph{starting} the software, ii) \emph{checking license}, iii) \emph{importing images} from local SSD before, and v) \emph{exporting poses} as XMP, vi) \emph{saving} the project to local SSD and vii) \emph{closing} the software after the iv) \emph{alignment} step.
In order to find a good balance between CPU and GPU compute power, all our Reality Capture experiments were run on cloud machine with 48 logical CPU cores (\mbox{Intel\textsuperscript{\textregistered}} Skylake, Xeon@\SI{2.00}{Ghz}), a single \mbox{NVIDIA\textsuperscript{\textregistered}} T4~\cite{nvidiat4} GPU and 312GB RAM.

\begin{table}[h!]
  \adjustbox{width=\linewidth}{
} \\
        \bottomrule
    \end{tabular}}
    \caption{\textbf{Tanks and Temples - SSIM ($\uparrow$).} We show the pose accuracy via view synthesis with Nerfacto~\cite{nerfstudio} as \mbox{\textbf{SSIM}~($\uparrow$)~\cite{wang2004image}}, and the reconstruction time. We color code results compared to COLMAP, \emph{default} and \emph{fast}, respectively: \TCBTableDarkGreen{$>0.05$ better}, \TCBTableLightGreen{within $\pm 0.05$}, \TCBTableYellow{$>0.05$ worse}, \TCBTableRed{$>0.1$ worse}. 
    $^{\dagger}$Method needs sequential inputs.}%
    \label{tab:supp:t2_ssim}%
\end{table}

\begin{table}[h!]
  \adjustbox{width=\linewidth}{
} \\
        \bottomrule
    \end{tabular}}
    \caption{\textbf{Tanks and Temples - LPIPS ($\downarrow$).} We show the pose accuracy via view synthesis with Nerfacto~\cite{nerfstudio} as \mbox{\textbf{LPIPS}~($\downarrow$)~\cite{zhang2018perceptual}}, and the reconstruction time. We color code results compared to COLMAP, \emph{default} and \emph{fast}, respectively: \TCBTableDarkGreen{$>0.05$ better (lower)}, \TCBTableLightGreen{within $\pm 0.05$}, \TCBTableYellow{$>0.05$ worse (higher)}, \TCBTableRed{$>0.1$ worse (higher)}. 
    $^{\dagger}$Method needs sequential inputs.}%
    \label{tab:supp:t2_lpips}%
\end{table}

\begin{table}[]
  \begin{minipage}{0.495\linewidth}
    \adjustbox{width=\linewidth}{
      \begin{tabular}{@{}lc||cccc@{}}
        &&&&&\\[-8pt] 
        \toprule
        & \multicolumn{1}{c||}{~Pseudo GT~}%
        & ~DROID-SLAM$^\dagger$~
        & ~BARF~
        & ~NoPe-NeRF$^\dagger$~
        &\multicolumn{1}{c}{~ACE0~}%
        \\%
        &\multicolumn{1}{c||}{(COLMAP)}%
        & \cite{teed2021droid}%
        & \cite{lin2021barf}%
        & \cite{bian2023nope}%
        &\multicolumn{1}{c}{(ours)}%
        \\%
        \hline
        Bicycle  & 0.68        &    0.20    & 0.22    & 0.24    & \textbf{0.47} \\
        Bonsai   & 0.89        &    0.18    & 0.31    & 0.36    & \textbf{0.83} \\
        Counter  & 0.82        &    0.25    & 0.26    & 0.27    & \textbf{0.77} \\
        Garden   & 0.86        &    0.38    & 0.27    & 0.26    & \textbf{0.77} \\
        Kitchen  & 0.87        &    0.28    & 0.34    & 0.33    & \textbf{0.80} \\
        Room     & 0.91        &    0.34    & 0.36    & 0.38    & \textbf{0.61} \\
        Stump    & 0.40        &    0.23    & 0.23    & 0.21    & \textbf{0.40} \\
        \hline
        Average  & 0.78        &    0.26    & 0.28    & 0.30    & \textbf{0.67} \\
    \end{tabular} 
    }
    \caption{%
      \textbf{Mip-NeRF 360 - SSIM ($\uparrow$).} Pose quality in \mbox{\textbf{SSIM}~($\uparrow$)~\cite{wang2004image}}, higher is better. %
      Best in \textbf{bold}. %
      $^\dagger$Method needs sequential inputs. %
    }%
    \label{tab:supp:mip_ssim}%
  \end{minipage}
  \hfill
  \begin{minipage}{0.495\linewidth}
    \adjustbox{width=\linewidth}{
      \begin{tabular}{@{}lc||cccc@{}}
        &&&&&\\[-8pt] 
        \toprule
        & \multicolumn{1}{c||}{~Pseudo GT~}%
        & ~DROID-SLAM$^\dagger$~
        & ~BARF~
        & ~NoPe-NeRF$^\dagger$~
        &\multicolumn{1}{c}{~ACE0~}%
        \\%
        &\multicolumn{1}{c||}{(COLMAP)}%
        & \cite{teed2021droid}%
        & \cite{lin2021barf}%
        & \cite{bian2023nope}%
        &\multicolumn{1}{c}{(ours)}%
        \\%
        \hline
        Bicycle  & 0.20        &    0.80    & 0.82    & 0.85    & \textbf{0.33} \\
        Bonsai   & 0.06        &    0.74    & 0.86    & 0.66    & \textbf{0.11} \\
        Counter  & 0.10        &    0.64    & 0.83    & 0.76    & \textbf{0.12} \\
        Garden   & 0.08        &    0.38    & 0.84    & 0.84    & \textbf{0.12} \\
        Kitchen  & 0.06        &    0.71    & 0.84    & 0.74    & \textbf{0.11} \\
        Room     & 0.05        &    0.85    & 0.83    & 0.59    & \textbf{0.43} \\
        Stump    & 0.45        &    0.85    & 0.79    & 0.81    & \textbf{0.36} \\
        \hline
        Average  & 0.14        &    0.71    & 0.83    & 0.75    & \textbf{0.23} \\
    \end{tabular}
    }
    \caption{%
      \textbf{Mip-NeRF 360 - LPIPS ($\downarrow$).} Pose quality in \mbox{\textbf{LPIPS}~($\downarrow$)~\cite{zhang2018perceptual}}, lower is better. %
      Best in \textbf{bold}. %
      $^\dagger$Method needs sequential inputs. %
    }%
    \label{tab:supp:mip_lpips}%
  \end{minipage}
\end{table}

\begin{table}
\centering
\scriptsize	
\begin{tabular}{|ll|}
\hline
random\_seed & 0\\
Mapper.min\_num\_matches & 15\\
Mapper.ignore\_watermarks & 0\\
Mapper.multiple\_models & 1\\
Mapper.max\_num\_models & 50\\
Mapper.max\_model\_overlap & 20\\
Mapper.min\_model\_size & 10\\
Mapper.init\_image\_id1 & -1\\
Mapper.init\_image\_id2 & -1\\
Mapper.init\_num\_trials & 200\\
Mapper.extract\_colors & 1\\
Mapper.num\_threads & -1\\
Mapper.min\_focal\_length\_ratio & 0.1 \\
Mapper.max\_focal\_length\_ratio & 10\\
Mapper.max\_extra\_param & 1\\
Mapper.ba\_refine\_focal\_length & 1\\
Mapper.ba\_refine\_principal\_point & 0\\
Mapper.ba\_refine\_extra\_params & 1\\
Mapper.ba\_min\_num\_residuals\_for\_multi\_threading & 50000\\
Mapper.ba\_local\_num\_images & 6\\
Mapper.ba\_local\_function\_tolerance & 0\\
Mapper.ba\_local\_max\_num\_iterations & 25\\
Mapper.ba\_global\_use\_pba & 0\\
Mapper.ba\_global\_pba\_gpu\_index & -1\\
\textbf{Mapper.ba\_global\_images\_ratio} & 1.1\\
\textbf{Mapper.ba\_global\_points\_ratio} & 1.1\\
Mapper.ba\_global\_images\_freq & 500\\
\textbf{Mapper.ba\_global\_points\_freq} & 250000\\
Mapper.ba\_global\_function\_tolerance & 0\\
\textbf{Mapper.ba\_global\_max\_num\_iterations} & 50\\
\textbf{Mapper.ba\_global\_max\_refinements} & 5\\
Mapper.ba\_global\_max\_refinement\_change & 0.0005\\
Mapper.ba\_local\_max\_refinements & 2\\
Mapper.ba\_local\_max\_refinement\_change & 0.001\\
Mapper.snapshot\_images\_freq & 0\\
Mapper.fix\_existing\_images & 0\\
Mapper.init\_min\_num\_inliers & 100\\
Mapper.init\_max\_error & 4\\
Mapper.init\_max\_forward\_motion & 0.95 \\
Mapper.init\_min\_tri\_angle & 16\\
Mapper.init\_max\_reg\_trials & 2\\
Mapper.abs\_pose\_max\_error & 12\\
Mapper.abs\_pose\_min\_num\_inliers & 30\\
Mapper.abs\_pose\_min\_inlier\_ratio & 0.25\\
Mapper.filter\_max\_reproj\_error & 4\\
Mapper.filter\_min\_tri\_angle & 1.5\\
Mapper.max\_reg\_trials & 3\\
Mapper.local\_ba\_min\_tri\_angle & 6\\
Mapper.tri\_max\_transitivity & 1\\
Mapper.tri\_create\_max\_angle\_error & 2\\
Mapper.tri\_continue\_max\_angle\_error & 2\\
Mapper.tri\_merge\_max\_reproj\_error & 4\\
Mapper.tri\_complete\_max\_reproj\_error & 4\\
Mapper.tri\_complete\_max\_transitivity & 5\\
Mapper.tri\_re\_max\_angle\_error & 5\\
Mapper.tri\_re\_min\_ratio & 0.2\\
Mapper.tri\_re\_max\_trials & 1\\
Mapper.tri\_min\_angle & 1.5\\
Mapper.tri\_ignore\_two\_view\_tracks & 1\\
\hline
\end{tabular}
\caption{COLMAP's \emph{default} incremental \texttt{mapper} options. \textbf{Bold} denotes options later changed in the \emph{fast} version, \cf, \Cref{tab:supp:colmap_mapper_fast}.}
\label{tab:colmap_mapper_default}
\end{table}

\begin{table}[]
\centering
\footnotesize
\begin{tabular}{|ll|}
\hline
  \textbf{Mapper.ba\_global\_images\_ratio} & \textbf{1.2} \\
 \textbf{Mapper.ba\_global\_points\_ratio} & \textbf{1.2} \\
  \textbf{Mapper.ba\_global\_points\_freq} & \textbf{200000} \\
  \textbf{Mapper.ba\_global\_max\_num\_iterations} & \textbf{20} \\
 \textbf{Mapper.ba\_global\_max\_refinements} & \textbf{3} \\
\hline
\end{tabular}
\caption{The incremental \texttt{mapper} options we used when running COLMAP \emph{fast}.
Other parameters were not changed, and can be seen in \Cref{tab:colmap_mapper_default}.
Parameters were recommended for large image collections of more than 1000 images by \cite{fastcolmap}.%
}
\label{tab:supp:colmap_mapper_fast}
\end{table}

\begin{table}[]
\centering
\footnotesize
    \begin{tabular}{|ll|}
    \hline
    random\_seed & 0 \\
    descriptor\_normalization & l1\_root \\
    ImageReader.mask\_path & - \\
    \textbf{ImageReader.camera\_model} & \textbf{SIMPLE\_PINHOLE} \\
    \textbf{ImageReader.single\_camera} & \textbf{1} \\
    ImageReader.single\_camera\_per\_folder & 0 \\
    ImageReader.single\_camera\_per\_image & 0\\
    ImageReader.existing\_camera\_id & -1 \\
    \textbf{ImageReader.camera\_params} & \textbf{f, cx, cy} from dataset \\
    ImageReader.default\_focal\_length\_factor & 1.2 \\
    ImageReader.camera\_mask\_path & - \\
    SiftExtraction.num\_threads & -1 \\
    SiftExtraction.use\_gpu & 1 \\
    SiftExtraction.gpu\_index & -1\\
    SiftExtraction.max\_image\_size & 3200\\
    SiftExtraction.max\_num\_features & 8192\\
    SiftExtraction.first\_octave & -1\\
    SiftExtraction.num\_octaves & 4\\
    SiftExtraction.octave\_resolution & 3\\
    SiftExtraction.peak\_threshold & $0.006\dot{6}$\\
    SiftExtraction.edge\_threshold & 10\\
    SiftExtraction.estimate\_affine\_shape & 0\\
    SiftExtraction.max\_num\_orientations & 2\\
    SiftExtraction.upright & 0\\
    SiftExtraction.domain\_size\_pooling & 0\\
    SiftExtraction.dsp\_min\_scale & $0.16\dot{6}$\\
    SiftExtraction.dsp\_max\_scale & 3\\
    SiftExtraction.dsp\_num\_scales & 10\\
    \hline
    \end{tabular}
    \caption{COLMAP's \texttt{feature\_extractor} options.
    \textbf{Bold} denotes options changed w.r.t. default.
    Ran on an \mbox{NVIDIA\textsuperscript{\textregistered}} V100\cite{nvidiav100} GPU.
    }
    \label{tab:colmap_feature_extractor}
\end{table}

\begin{table}[]
\centering
\footnotesize
     \setlength{\tabcolsep}{1pt}%
    \begin{tabular}{|ll|}
    \hline
random\_seed & 0\\
SiftMatching.num\_threads & -1\\
SiftMatching.use\_gpu & 1\\
SiftMatching.gpu\_index & -1\\
SiftMatching.max\_ratio & 0.8\\
SiftMatching.max\_distance & 0.7\\
SiftMatching.cross\_check & 1\\
SiftMatching.max\_error & 4\\
SiftMatching.max\_num\_matches & 32768\\
SiftMatching.confidence & 0.999\\
SiftMatching.max\_num\_trials & 10000\\
SiftMatching.min\_inlier\_ratio & 0.25\\
SiftMatching.min\_num\_inliers & 15\\
SiftMatching.multiple\_models & 0\\
SiftMatching.guided\_matching & 0\\
SiftMatching.planar\_scene & 0\\
SiftMatching.compute\_relative\_pose & 0\\
VocabTreeMatching.num\_images & 100\\
VocabTreeMatching.num\_nearest\_neighbors & 5\\
VocabTreeMatching.num\_checks & 256\\
{\scriptsize \text{VocabTreeMatching.num\_images\_after\_verification}} & 0\\
VocabTreeMatching.max\_num\_features & -1\\
VocabTreeMatching.vocab\_tree\_path & \multicolumn{1}{l|}{\begin{tabular}[c]{@{}l@{}}\tiny https://demuc.de/colmap/\\ \tiny vocab\_tree\_flickr100K\_words256K.bin\end{tabular}} \\
    \hline
    \end{tabular}
    \caption{COLMAP's default \texttt{vocab\_tree\_matcher} options used. Ran on an \mbox{NVIDIA\textsuperscript{\textregistered}} V100\cite{nvidiav100} GPU.}
    \label{tab:supp:colmap_feature_matcher}
\end{table}

\begin{table}[]
    \centering
    \footnotesize
    \setlength{\tabcolsep}{1pt}%
    \begin{tabular}{|@{}ll@{}|}
      \hline
      \textbf{Mapper.ba\_global\_images\_ratio} & \textbf{1.2} \\
      \textbf{Mapper.ba\_global\_points\_ratio} & \textbf{1.2} \\
      \textbf{Mapper.ba\_global\_max\_num\_iterations} & \textbf{20} \\
      \textbf{Mapper.ba\_global\_max\_refinements} & \textbf{3} \\
      \textbf{Mapper.ba\_global\_points\_freq} & \textbf{200\hspace{1pt}000} \\
      \hline
    \end{tabular}
    \caption{The \texttt{image\_registrator} options used for COLMAP in the \colmapreloc step.
      \textbf{Bold} denotes options changed w.r.t. default.
      Parameters set to match \Cref{tab:supp:colmap_mapper_fast}, as recommended for large image collections of more than \num{1000} images by \cite{fastcolmap}.%
    }
    \label{tab:supp:colmap_image_registrator}
\end{table}

\begin{table}[]
    \centering
    \footnotesize
    \setlength{\tabcolsep}{1pt}%
    \begin{tabular}{|@{}ll@{}|}
      \hline
        \textbf{BundleAdjustment.max\_num\_iterations} & \textbf{20} \\
      \hline
    \end{tabular}
    \caption{The \texttt{bundle\_adjuster} options used for COLMAP in the \colmapreloc step.
      \textbf{Bold} denotes options changed w.r.t. default.
      Parameters set to match \Cref{tab:supp:colmap_mapper_fast}, as recommended for large image collections of more than \num{1000} images by \cite{fastcolmap}.%
    }
    \label{tab:supp:colmap_bundle_adjuster}
\end{table}

%
%
\bibliographystyle{splncs04}
\bibliography{main}

\begin{thebibliography}{100}
\providecommand{\url}[1]{\texttt{#1}}
\providecommand{\urlprefix}{URL }
\providecommand{\doi}[1]{https://doi.org/#1}

\bibitem{agarwal2011building}
Agarwal, S., Furukawa, Y., Snavely, N., Simon, I., Curless, B., Seitz, S.M., Szeliski, R.: Building {R}ome in a day. ACM TOG  (2011)

\bibitem{agarwal2010bundle}
Agarwal, S., Snavely, N., Seitz, S.M., Szeliski, R.: Bundle adjustment in the large. In: ECCV (2010)

\bibitem{arnold2022map}
Arnold, E., Wynn, J., Vicente, S., Garcia-Hernando, G., Monszpart, A., Prisacariu, V.A., Turmukhambetov, D., Brachmann, E.: Map-free visual relocalization: Metric pose relative to a single image. In: ECCV (2022)

\bibitem{balntas2018relocnet}
Balntas, V., Li, S., Prisacariu, V.A.: Reloc{N}et: Continuous metric learning relocalisation using neural nets. In: ECCV (2018)

\bibitem{barron2022mipnerf360}
Barron, J.T., Mildenhall, B., Verbin, D., Srinivasan, P.P., Hedman, P.: {Mip-NeRF 360}: Unbounded anti-aliased neural radiance fields. In: CVPR (2022)

\bibitem{beardsley1997sequential}
Beardsley, P.A., Zisserman, A., Murray, D.W.: Sequential updating of projective and affine structure from motion. IJCV  (1997)

\bibitem{bhat2023zoedepth}
Bhat, S.F., Birkl, R., Wofk, D., Wonka, P., M{\"u}ller, M.: Zoe{D}epth: Zero-shot transfer by combining relative and metric depth. arXiv  (2023)

\bibitem{bhowmick2015divide}
Bhowmick, B., Patra, S., Chatterjee, A., Govindu, V.M., Banerjee, S.: Divide and conquer: Efficient large-scale structure from motion using graph partitioning. In: ACCV (2015)

\bibitem{bhowmick2017divide}
Bhowmick, B., Patra, S., Chatterjee, A., Govindu, V.M., Banerjee, S.: Divide and conquer: A hierarchical approach to large-scale structure-from-motion. {CVIU}  (2017)

\bibitem{bian2023nope}
Bian, W., Wang, Z., Li, K., Bian, J.W., Prisacariu, V.A.: {NoPe-NeRF}: Optimising neural radiance field with no pose prior. In: CVPR (2023)

\bibitem{Bloesch2018codeslam}
Bloesch, M., Czarnowski, J., Clark, R., Leutenegger, S., Davison, A.J.: {CodeSLAM} --- {L}earning a compact, optimisable representation for dense visual {SLAM}. In: CVPR (2018)

\bibitem{brachmann2023accelerated}
Brachmann, E., Cavallari, T., Prisacariu, V.A.: Accelerated coordinate encoding: Learning to relocalize in minutes using {RGB} and poses. In: CVPR (2023)

\bibitem{brachmann2021limits}
Brachmann, E., Humenberger, M., Rother, C., Sattler, T.: On the limits of pseudo ground truth in visual camera re-localisation. In: ICCV (2021)

\bibitem{brachmann2017dsac}
Brachmann, E., Krull, A., Nowozin, S., Shotton, J., Michel, F., Gumhold, S., Rother, C.: {DSAC}-differentiable {RANSAC} for camera localization. In: CVPR (2017)

\bibitem{brachmann2018dsacpp}
Brachmann, E., Rother, C.: Learning less is more-{6D} camera localization via {3D} surface regression. In: CVPR (2018)

\bibitem{brachmann2019expert}
Brachmann, E., Rother, C.: Expert sample consensus applied to camera re-localization. In: ICCV (2019)

\bibitem{brachmann2021dsacstar}
Brachmann, E., Rother, C.: Visual camera re-localization from {RGB} and {RGB-D} images using {DSAC}. IEEE TPAMI  (2021)

\bibitem{bregier2021deepregression}
Br{\'e}gier, R.: Deep regression on manifolds: a {3D} rotation case study. In: 3DV (2021)

\bibitem{brown1976bundle}
Brown, D.: The bundle adjustment-progress and prospect. In: {Congr. of the Int. Soc. for Photogr.} (1976)

\bibitem{brown2005unsupervised}
Brown, M., Lowe, D.G.: Unsupervised {3D} object recognition and reconstruction in unordered datasets. In: {3DIM} (2005)

\bibitem{carlone2015initialization}
Carlone, L., Tron, R., Daniilidis, K., Dellaert, F.: Initialization techniques for {3D SLAM}: A survey on rotation estimation and its use in pose graph optimization. In: {ICRA} (2015)

\bibitem{cavallari2019let}
Cavallari, T., Bertinetto, L., Mukhoti, J., Torr, P.H., Golodetz, S.: Let's take this online: Adapting scene coordinate regression network predictions for online {RGB-D} camera relocalisation. In: 3DV (2019)

\bibitem{cavallari2017fly}
Cavallari, T., Golodetz, S., Lord, N.A., Valentin, J., Di~Stefano, L., Torr, P.H.: On-the-fly adaptation of regression forests for online camera relocalisation. In: CVPR (2017)

\bibitem{chen2023refinement}
Chen, S., Bhalgat, Y., Li, X., Bian, J., Li, K., Wang, Z., Prisacariu, V.A.: Neural refinement for absolute pose regression with feature synthesis. In: CVPR (2024)

\bibitem{chen2022dfnet}
Chen, S., Li, X., Wang, Z., Prisacariu, V.: {DFN}et: {E}nhance absolute pose regression with direct feature matching. In: ECCV (2022)

\bibitem{chen21}
Chen, S., Wang, Z., Prisacariu, V.: {Direct-PoseNet}: Absolute pose regression with photometric consistency. In: 3DV (2021)

\bibitem{cheng2023lunerf}
Cheng, Z., Esteves, C., Jampani, V., Kar, A., Maji, S., Makadia, A.: {LU-NeRF}: Scene and pose estimation by synchronizing local unposed {NeRFs}. In: ICCV (2023)

\bibitem{crandall2011discrete}
Crandall, D., Owens, A., Snavely, N., Huttenlocher, D.: Discrete-continuous optimization for large-scale structure from motion. In: CVPR (2011)

\bibitem{dai2017scannet}
Dai, A., Chang, A.X., Savva, M., Halber, M., Funkhouser, T., Nie{\ss}ner, M.: Scan{N}et: Richly-annotated {3D} reconstructions of indoor scenes. In: CVPR (2017)

\bibitem{davison2003real}
Davison, A.J.: Real-time simultaneous localisation and mapping with a single camera. In: ICCV (2003)

\bibitem{detone2018superpoint}
DeTone, D., Malisiewicz, T., Rabinovich, A.: Super{P}oint: Self-supervised interest point detection and description. In: CVPRW (2018)

\bibitem{ding2019camnet}
Ding, M., Wang, Z., Sun, J., Shi, J., Luo, P.: Cam{N}et: Coarse-to-fine retrieval for camera re-localization. In: ICCV (2019)

\bibitem{dusmanu2019d2}
Dusmanu, M., Rocco, I., Pajdla, T., Pollefeys, M., Sivic, J., Torii, A., Sattler, T.: D2-net: A trainable {CNN} for joint description and detection of local features. In: CVPR (2019)

\bibitem{ransac}
Fischler, M.A., Bolles, R.C.: Random sample consensus: a paradigm for model fitting with applications to image analysis and automated cartography. Comm. of the ACM  (1981)

\bibitem{pnp}
Gao, X.S., Hou, X.R., Tang, J., Cheng, H.F.: Complete solution classification for the perspective-three-point problem. IEEE TPAMI  (2003)

\bibitem{gherardi2010improving}
Gherardi, R., Farenzena, M., Fusiello, A.: Improving the efficiency of hierarchical structure-and-motion. In: CVPR (2010)

\bibitem{govindu2001combining}
Govindu, V.M.: Combining two-view constraints for motion estimation. In: CVPR (2001)

\bibitem{govindu2004lie}
Govindu, V.M.: Lie-algebraic averaging for globally consistent motion estimation. In: CVPR (2004)

\bibitem{hartley2013rotation}
Hartley, R., Trumpf, J., Dai, Y., Li, H.: Rotation averaging. IJCV  (2013)

\bibitem{hartley2003multiple}
Hartley, R., Zisserman, A.: Multiple view geometry in computer vision. Cambridge University Press (2003)

\bibitem{he2024detectorfree}
He, X., Sun, J., Wang, Y., Peng, S., Huang, Q., Bao, H., Zhou, X.: Detector-free structure from motion. In: CVPR (2024)

\bibitem{heinly2015reconstructing}
Heinly, J., Sch\"{o}nberger, J.L., Dunn, E., Frahm, J.M.: Reconstructing the {W}orld in six days. In: CVPR (2015)

\bibitem{humenberger2022investigating}
Humenberger, M., Cabon, Y., Pion, N., Weinzaepfel, P., Lee, D., Gu{\'e}rin, N., Sattler, T., Csurka, G.: Investigating the role of image retrieval for visual localization: An exhaustive benchmark. IJCV  (2022)

\bibitem{izadi2011kinectfusion}
Izadi, S., Kim, D., Hilliges, O., Molyneaux, D., Newcombe, R., Kohli, P., Shotton, J., Hodges, S., Freeman, D., Davison, A.J., Fitzgibbon, A.: Kinect{F}usion: real-time {3D} reconstruction and interaction using a moving depth camera. In: {UIST} (2011)

\bibitem{jeong2021self}
Jeong, Y., Ahn, S., Choy, C., Anandkumar, A., Cho, M., Park, J.: Self-calibrating neural radiance fields. In: ICCV (2021)

\bibitem{jin2021image}
Jin, Y., Mishkin, D., Mishchuk, A., Matas, J., Fua, P., Yi, K.M., Trulls, E.: Image matching across wide baselines: From paper to practice. IJCV  (2021)

\bibitem{kabsch1976solution}
Kabsch, W.: A solution for the best rotation to relate two sets of vectors. Acta Crystallographica Section A: Crystal Physics, Diffraction, Theoretical and General Crystallography  \textbf{32}(5),  922--923 (1976)

\bibitem{kendall2015posenet}
Kendall, A., Grimes, M., Cipolla, R.: Pose{N}et: A convolutional network for real-time {6-DoF} camera relocalization. In: ICCV (2015)

\bibitem{kerbl3Dgaussians}
Kerbl, B., Kopanas, G., Leimk{\"u}hler, T., Drettakis, G.: {3D} gaussian splatting for real-time radiance field rendering. ACM TOG  (2023)

\bibitem{tanksandtemples}
Knapitsch, A., Park, J., Zhou, Q.Y., Koltun, V.: Tanks and {T}emples: Benchmarking large-scale scene reconstruction. ACM TOG  (2017)

\bibitem{kraus1993photogrammetry}
Kraus, K.: Photogrammetry. No.~v. 1 in Photogrammetry, Ferdinand Dummlers Verlag (1993)

\bibitem{laskar2017camera}
Laskar, Z., Melekhov, I., Kalia, S., Kannala, J.: Camera relocalization by computing pairwise relative poses using convolutional neural network. In: {ICCV Workshops} (2017)

\bibitem{li2020hierarchical}
Li, X., Wang, S., Zhao, Y., Verbeek, J., Kannala, J.: Hierarchical scene coordinate classification and regression for visual localization. In: CVPR (2020)

\bibitem{lin2024relposepp}
Lin, A., Zhang, J.Y., Ramanan, D., Tulsiani, S.: Relpose++: Recovering 6d poses from sparse-view observations. In: 3DV (2024)

\bibitem{lin2021barf}
Lin, C.H., Ma, W.C., Torralba, A., Lucey, S.: {BARF}: Bundle-adjusting neural radiance fields. In: ICCV (2021)

\bibitem{lin2023icrapnerf}
Lin, Y., M{\"u}ller, T., Tremblay, J., Wen, B., Tyree, S., Evans, A., Vela, P.A., Birchfield, S.: Parallel inversion of neural radiance fields for robust pose estimation. In: {ICRA} (2023)

\bibitem{lindenberger2023lightglue}
Lindenberger, P., Sarlin, P.E., Pollefeys, M.: Light{G}lue: Local feature matching at light speed. In: ICCV (2023)

\bibitem{liu2019planercnn}
Liu, C., Kim, K., Gu, J., Furukawa, Y., Kautz, J.: Plane{RCNN}: {3D} plane detection and reconstruction from a single image. In: CVPR (2019)

\bibitem{loshchilov2019decoupled}
Loshchilov, I., Hutter, F.: Decoupled weight decay regularization. In: ICLR (2019)

\bibitem{lowe2004distinctive}
Lowe, D.G.: Distinctive image features from scale-invariant keypoints. IJCV  (2004)

\bibitem{martinec2007robust}
Martinec, D., Pajdla, T.: Robust rotation and translation estimation in multiview reconstruction. In: CVPR (2007)

\bibitem{meng2021gnerf}
Meng, Q., Chen, A., Luo, H., Wu, M., Su, H., Xu, L., He, X., Yu, J.: {G}{N}e{R}{F}: {G}{A}{N}-based {N}eural {R}adiance {F}ield without {P}osed {C}amera. In: ICCV (2021)

\bibitem{mildenhall2020nerf}
Mildenhall, B., Srinivasan, P.P., Tancik, M., Barron, J.T., Ramamoorthi, R., Ng, R.: {NeRF}: Representing scenes as neural radiance fields for view synthesis. In: ECCV (2020)

\bibitem{moreau2023crossfire}
Moreau, A., Piasco, N., Bennehar, M., Tsishkou, D., Stanciulescu, B., de~La~Fortelle, A.: {CROSSFIRE}: Camera relocalization on self-supervised features from an implicit representation. ICCV  (2023)

\bibitem{mueller2022instant}
M\"uller, T., Evans, A., Schied, C., Keller, A.: Instant neural graphics primitives with a multiresolution hash encoding. ACM TOG  (2022)

\bibitem{newcombe2011kinectfusion}
Newcombe, R., Izadi, S., Hilliges, O., Molyneaux, D., Kim, D., Davison, A.J., Kohi, P., Shotton, J., Hodges, S., Fitzgibbon, A.: Kinect{F}usion: Real-time dense surface mapping and tracking. In: {ISMAR} (2011)

\bibitem{nister2004visual}
Nist{\'e}r, D., Naroditsky, O., Bergen, J.: Visual odometry. In: CVPR (2004)

\bibitem{nvidiav100}
{NVIDIA\textsuperscript{\textregistered}}: {NVIDIA V100 TENSOR CORE GPU}. \url{https://www.nvidia.com/en-us/data-center/tesla-v100/} (2017), [acessed March/14/2024]

\bibitem{nvidiat4}
{NVIDIA\textsuperscript{\textregistered}}: {NVIDIA T4}. \url{https://www.nvidia.com/en-us/data-center/tesla-t4/} (2018), [acessed March/14/2024]

\bibitem{pollefeys2000automated}
Pollefeys, M., Koch, R., Vergauwen, M., Van~Gool, L.: Automated reconstruction of {3D} scenes from sequences of images. J. of Photogr. and Rem. Sens.  (2000)

\bibitem{rau2020predicting}
Rau, A., Garcia-Hernando, G., Stoyanov, D., Brostow, G.J., Turmukhambetov, D.: Predicting visual overlap of images through interpretable non-metric box embeddings. In: ECCV (2020)

\bibitem{reality-capture}
Reality, C.: {Reality Capture}. \url{https://www.capturingreality.com/realitycapture} (2016), [accessed 15-Nov-2023]

\bibitem{reizenstein2021common}
Reizenstein, J., Shapovalov, R., Henzler, P., Sbordone, L., Labatut, P., Novotny, D.: Common objects in {3D}: Large-scale learning and evaluation of real-life {3D} category reconstruction. In: ICCV (2021)

\bibitem{sarlin2019coarse}
Sarlin, P.E., Cadena, C., Siegwart, R., Dymczyk, M.: From coarse to fine: Robust hierarchical localization at large scale. In: CVPR (2019)

\bibitem{sarlin2020superglue}
Sarlin, P.E., DeTone, D., Malisiewicz, T., Rabinovich, A.: Super{G}lue: Learning feature matching with graph neural networks. In: CVPR (2020)

\bibitem{sarlin2021back}
Sarlin, P.E., Unagar, A., Larsson, M., Germain, H., Toft, C., Larsson, V., Pollefeys, M., Lepetit, V., Hammarstrand, L., Kahl, F., Sattler, T.: Back to the feature: Learning robust camera localization from pixels to pose. In: CVPR (2021)

\bibitem{sattler2011fast}
Sattler, T., Leibe, B., Kobbelt, L.: Fast image-based localization using direct {2D-to-3D} matching. In: ICCV (2011)

\bibitem{sattler2012improving}
Sattler, T., Leibe, B., Kobbelt, L.: Improving image-based localization by active correspondence search. In: ECCV (2012)

\bibitem{sattler2016efficient}
Sattler, T., Leibe, B., Kobbelt, L.: Efficient \& effective prioritized matching for large-scale image-based localization. IEEE TPAMI  (2016)

\bibitem{sattler2017large}
Sattler, T., Torii, A., Sivic, J., Pollefeys, M., Taira, H., Okutomi, M., Pajdla, T.: Are large-scale {3D} models really necessary for accurate visual localization? In: CVPR (2017)

\bibitem{sattler2019understanding}
Sattler, T., Zhou, Q., Pollefeys, M., Leal-Taixe, L.: Understanding the limitations of {CNN}-based absolute camera pose regression. In: CVPR (2019)

\bibitem{schaffalitzky2002multi}
Schaffalitzky, F., Zisserman, A.: Multi-view matching for unordered image sets, or “{H}ow do i organize my holiday snaps?”. In: ECCV (2002)

\bibitem{fastcolmap}
Sch\"{o}nberger, J.L.: {Colmap Github Issues}. \url{https://github.com/colmap/colmap/issues/116\#issuecomment-298926277} (2017), [accessed Nov/15/2023]

\bibitem{schoenberger2016sfm}
Sch\"{o}nberger, J.L., Frahm, J.M.: Structure-from-motion revisited. In: CVPR (2016)

\bibitem{shotton2013scene}
Shotton, J., Glocker, B., Zach, C., Izadi, S., Criminisi, A., Fitzgibbon, A.: Scene coordinate regression forests for camera relocalization in {RGB-D} images. In: CVPR (2013)

\bibitem{sinha2023sparsepose}
Sinha, S., Zhang, J.Y., Tagliasacchi, A., Gilitschenski, I., Lindell, D.B.: Sparse{P}ose: Sparse-view camera pose regression and refinement. In: CVPR (2023)

\bibitem{smith2019super}
Smith, L.N., Topin, N.: Super-convergence: Very fast training of neural networks using large learning rates. In: Arti. Intell. and Mach. Learn. for Multi-Domain Operations Appli. (2019)

\bibitem{snavely2006photo}
Snavely, N., Seitz, S.M., Szeliski, R.: Photo tourism: exploring photo collections in {3D}. ACM TOG  (2006)

\bibitem{snavely2008modeling}
Snavely, N., Seitz, S.M., Szeliski, R.: Modeling the world from internet photo collections. IJCV  (2008)

\bibitem{snavely2008skeletal}
Snavely, N., Seitz, S.M., Szeliski, R.: Skeletal graphs for efficient structure from motion. In: CVPR (2008)

\bibitem{sun2021loftr}
Sun, J., Shen, Z., Wang, Y., Bao, H., Zhou, X.: {LoFTR}: Detector-free local feature matching with transformers. In: CVPR (2021)

\bibitem{szeliski1994recovering}
Szeliski, R., Kang, S.B.: Recovering {3D} shape and motion from image streams using nonlinear least squares. J. of Vis. Com. and Image Repr.  (1994)

\bibitem{nerfstudio}
Tancik, M., Weber, E., Ng, E., Li, R., Yi, B., Kerr, J., Wang, T., Kristoffersen, A., Austin, J., Salahi, K., Ahuja, A., McAllister, D., Kanazawa, A.: Nerfstudio: A modular framework for neural radiance field development. In: ACM TOG (2023)

\bibitem{teed2021droid}
Teed, Z., Deng, J.: {DROID-SLAM}: Deep visual {SLAM} for monocular, stereo, and {RGB-D} cameras. In: NeurIPS (2021)

\bibitem{toldo2015hierarchical}
Toldo, R., Gherardi, R., Farenzena, M., Fusiello, A.: Hierarchical structure-and-motion recovery from uncalibrated images. {CVIU}  (2015)

\bibitem{triggs2000bundle}
Triggs, B., McLauchlan, P.F., Hartley, R.I., Fitzgibbon, A.W.: Bundle adjustment—a modern synthesis. In: {Int. Worksh. on Vis. Alg.} (2000)

\bibitem{turkoglu2021visual}
T{\"{u}}rko\u{g}lu, M.{\"{O}}., Brachmann, E., Schindler, K., Brostow, G., Monszpart, A.: Visual camera re-localization using graph neural networks and relative pose supervision. In: 3DV (2021)

\bibitem{ulyanov2018deep}
Ulyanov, D., Vedaldi, A., Lempitsky, V.: Deep image prior. In: CVPR (2018)

\bibitem{ummenhofer2017demon}
Ummenhofer, B., Zhou, H., Uhrig, J., Mayer, N., Ilg, E., Dosovitskiy, A., Brox, T.: {DeMoN}: Depth and motion network for learning monocular stereo. In: CVPR (2017)

\bibitem{waechter2017rephotography}
Waechter, M., Beljan, M., Fuhrmann, S., Moehrle, N., Kopf, J., Goesele, M.: Virtual rephotography: Novel view prediction error for 3d reconstruction. ACM TOG  (2017)

\bibitem{wang2024dust3r}
Wang, S., Leroy, V., Cabon, Y., Chidlovskii, B., Revaud, J.: {DUSt3R}: Geometric {3D} vision made easy. In: CVPR (2024)

\bibitem{wang2004image}
Wang, Z., Bovik, A.C., Sheikh, H.R., Simoncelli, E.P.: Image quality assessment: from error visibility to structural similarity. IEEE TIP  (2004)

\bibitem{wang2021nerf}
Wang, Z., Wu, S., Xie, W., Chen, M., Prisacariu, V.A.: {NeRF}{-}{-}: Neural radiance fields without known camera parameters. arXiv  (2021)

\bibitem{wei2020deepsfm}
Wei, X., Zhang, Y., Li, Z., Fu, Y., Xue, X.: {DeepSFM}: Structure from motion via deep bundle adjustment. In: ECCV (2020)

\bibitem{wu2013towards}
Wu, C.: Towards linear-time incremental structure from motion. In: 3DV (2013)

\bibitem{xia2022sinerf}
Xia, Y., Tang, H., Timofte, R., Van~Gool, L.: {SiNeRF}: Sinusoidal neural radiance fields for joint pose estimation and scene reconstruction. In: BMVC (2022)

\bibitem{yen2021inerf}
Yen-Chen, L., Florence, P., Barron, J.T., Rodriguez, A., Isola, P., Lin, T.Y.: i{N}e{RF}: Inverting neural radiance fields for pose estimation. In: {IROS} (2021)

\bibitem{zhang2024raydiffusion}
Zhang, J.Y., Lin, A., Kumar, M., Yang, T.H., Ramanan, D., Tulsiani, S.: Cameras as rays: Pose estimation via ray diffusion. In: ICLR (2024)

\bibitem{zhang2018perceptual}
Zhang, R., Isola, P., Efros, A.A., Shechtman, E., Wang, O.: The unreasonable effectiveness of deep features as a perceptual metric. In: CVPR (2018)

\bibitem{zhang2006image}
Zhang, W., Kosecka, J.: Image based localization in urban environments. In: {3DPVT} (2006)

\bibitem{zhou2020learn}
Zhou, Q., Sattler, T., Pollefeys, M., Leal-Taixe, L.: To learn or not to learn: Visual localization from essential matrices. In: {ICRA} (2020)

\bibitem{Zhou2019Continuity}
Zhou, Y., Barnes, C., Jingwan, L., Jimei, Y., Hao, L.: On the continuity of rotation representations in neural networks. In: CVPR (2019)

\end{thebibliography}

\end{document}